\documentclass[11pt]{article}
\usepackage[preprint]{acl}
\usepackage{times}
\usepackage{latexsym}
\usepackage[T1]{fontenc}
\usepackage[utf8]{inputenc}
\usepackage{microtype}
\usepackage{graphicx}
\usepackage{booktabs}
\usepackage{array}
\usepackage{amsmath,amssymb}
\usepackage{xspace}
\usepackage{xcolor}
\usepackage[capitalise,noabbrev]{cleveref}
\newcommand{\sfgl}{SFG-L\xspace}
\newcommand{\dfedavg}{DFedAvg\xspace}
\title{When Does Freshness Weighting Help\\Decentralised Language-Model Training?}
\author{Qiyuan Wang \quad Ke Xiao \quad Christos Anagnostopoulos \\
  School of Computing Science, University of Glasgow}
\hypersetup{
  pdftitle={When Does Freshness Weighting Help Decentralised Language-Model Training?},
  pdfauthor={Qiyuan Wang, Ke Xiao, Christos Anagnostopoulos}
}
\begin{document}
\maketitle

\begin{abstract}
Clients fine-tuning a language model together can miss different rounds of training. We study the lag-based rule of Self-inclusive Freshness-weighted Gossip (SFG). This rule, \sfgl, discounts models by an inherited count of missed rounds. We fine-tuned Qwen2.5-7B across 100 clients with different mixtures of eight tasks. At validation-selected learning rates, the estimated gains in task-balanced accuracy over ordinary averaging were 0.72 percentage points under replayed smartphone availability and 0.09 under synthetic outages. At matched rates, the gains were 0.72 and 0.74 points at the lower rate and 0.63 and 0.94 at the higher rate. Neither paired comparison detected a difference in gain between schedules. Under replayed availability at the lower rate, WinoGrande accounted for 80\% of the gain. Ordinary averaging retained 53\% of the WinoGrande improvement from outage-free fine-tuning and \sfgl retained 87\%. Its advantage was 4.23 points on 767 additional items. Under synthetic outages at the higher rate, WinoGrande accounted for 29\% of the gain and HellaSwag improved almost as much. Tripling the learning rate helped both rules by 0.41 to 0.85 points and raised WikiText-103 loss. In GPT-124M pretraining, \sfgl lowered perplexity by 13.5\% after 900 rounds. From mature checkpoints the gain depended on the learning rate.
\end{abstract}

\section{Introduction}\label{sec:intro}

Clients that fine-tune a language model together can reach the same communication round with different training histories. While one client is unavailable, its neighbours may continue adapting their models on tasks that are rare in its own data. When the client returns, ordinary averaging gives its adapter the same weight as each available neighbour's. We investigate whether weighting adapters by a lag measure that tracks missed rounds and propagates through model exchanges helps clients retain the task improvements seen under uninterrupted fine-tuning. We ask which tasks account for the gain and whether the apparent difference between availability schedules persists when both rules use the same learning rate. \Cref{tab:takeaways} summarises the findings.

\begin{table*}[t]
\centering\small
\begin{tabular}{@{}>{\raggedright\arraybackslash}p{0.25\textwidth}@{\hspace{0.025\textwidth}}>{\raggedright\arraybackslash}p{0.725\textwidth}@{}}
\toprule
\textbf{Question} & \textbf{What the experiments showed} \\
\midrule
Does the Qwen gain depend on the outage schedule? &
At a shared learning rate of $10^{-4}$, \sfgl improved task-balanced accuracy on all five schedules. The gain under replayed smartphone availability (FLASH) replicated on three additional seeds. The schedule comparisons could not establish which properties of availability affected the gain (\cref{sec:qwen-avg,sec:qwen-schedule}). \\
\addlinespace
How does learning-rate selection change the comparison? &
When validation selected a rate for each rule, the gain was 0.72 points under FLASH and 0.09 under synthetic outages. At either matched rate we detected no difference in gain between schedules. Tripling the baseline rate left a 0.22-point advantage under FLASH (\cref{sec:qwen-lr}). \\
\addlinespace
Which tasks drive the average? &
At $10^{-4}$ under FLASH, WinoGrande accounted for 80\% of the gain and the other seven tasks gained 0.16 points on average. Under synthetic outages at $3\times10^{-4}$, WinoGrande accounted for 29\% and HellaSwag gained almost as much as WinoGrande (\cref{sec:qwen-tasks}). \\
\addlinespace
Which WinoGrande answers did clients retain? &
On items that outage-free fine-tuning corrected, accuracy under FLASH at $10^{-4}$ was 78.4\% with \sfgl and 51.9\% with ordinary averaging. On 767 additional WinoGrande items, \sfgl led by 4.23 points (\cref{sec:qwen-tasks}). \\
\addlinespace
What did higher task accuracy cost? &
At matched rates, \sfgl had higher task accuracy and higher WikiText-103 loss in all four comparisons. The interval for the loss difference excluded zero in three. Tripling the rate also improved task accuracy for both rules under outages and raised loss on this corpus for both (\cref{sec:qwen-other,sec:qwen-lr}). \\
\addlinespace
How long did the GPT pretraining gain last? &
From random initialisation, the reduction in perplexity narrowed from 19.0\% at 300 rounds to 13.5\% at 900. From the most mature checkpoint under synthetic outages, it was 2.0\% at the baseline's rate. We detected no advantage when each rule used its own selected rate (\cref{sec:gpt}). \\
\bottomrule
\end{tabular}
\caption{\textbf{Main takeaways.} Gains compare \sfgl with ordinary averaging. Qwen2.5-7B accuracy differences are in percentage points. GPT-124M perplexity reductions are relative percentages.}
\label{tab:takeaways}
\end{table*}

\paragraph{The rule and prior work.} In decentralised training there is no server: each client trains on its own data and averages its model with those of its neighbours \citep{lian2017dpsgd,sun2023dfedavg}. Methods for stale contributions usually discount what a client or server \emph{receives} by its age \citep{xie2019fedasync,nguyen2022fedbuff}. \citet{wang2026sfg} observed that after an absence the stalest model in the average is usually the receiving client's own. They proposed Self-inclusive Freshness-weighted Gossip (SFG), which also discounts the receiver's model when it is stale. We use their lag-based rule (\sfgl). Its \emph{inherited lag} counts the rounds a client misses and incorporates the lags of the models it averages. That work evaluated it on image classification, with a small convolutional network and a pretrained ResNet-18. Language models are increasingly trained and adapted across many participants \citep{ye2024openfedllm,douillard2023diloco,sani2024photon}, and they raise questions that image classification does not: what the rule preserves when clients adapt a pretrained model to different language tasks, whether an advantage in pretraining lasts as training continues, and how it depends on the maturity of the starting checkpoint.

\paragraph{What we do.} The main study fine-tunes Qwen2.5-7B with LoRA across 100 clients that each hold a different mixture of eight multiple-choice tasks (\cref{sec:qwen}). Clients are available according to a replay of recorded smartphone availability or a synthetic schedule with a similar share of rounds offline; three further schedules separate properties of the replay. A complementary study pretrains a GPT-124M model across 16 clients and varies the length of training and the checkpoint from which decentralised training starts (\cref{sec:gpt}). All experiments are simulations on server GPUs.

We use three terms for three different things. \emph{Unequal training histories} describe the setting. \emph{Weighting by inherited lag} is the intervention. \emph{Preservation of measured task gains} is the outcome. Lag does not measure task accuracy or what a client has learned; two equally fresh clients can have learned different things. We therefore ask whether accounting for participation history helps clients retain measured task gains, and make no claim that the rule aligns clients' progress.

\section{Setting and measurement}\label{sec:setting}

\paragraph{Rounds.} $N$ clients sit on a fixed graph and train in rounds. In each round a client is available or not. An available client first averages its model with the models of its available neighbours and then takes $K$ local optimiser steps on its own data. An unavailable client does neither and keeps its model. \emph{Ordinary averaging} gives every model in a client's average, its own included, the same weight; this is \dfedavg \citep{sun2023dfedavg}.

\paragraph{Inherited lag.} Each client $i$ carries one number $\ell_i$, initially zero. If the client is unavailable in a round, $\ell_i$ grows by one. If it is available and averages candidate models $j$ (its neighbours' and its own) with weights $w_{ij}$, then
\begin{equation}
\ell_i \leftarrow \textstyle\sum_j w_{ij}\,\ell_j .
\end{equation}
Lag therefore accumulates during absence and propagates through averaging: a client that was always available acquires lag by averaging with a lagging neighbour. \sfgl \citep{wang2026sfg} sets $w_{ij}\propto\gamma^{\ell_j-\ell_{\min}}$ with $\gamma=0.5$, where $\ell_{\min}$ is the smallest lag among the candidates. With equal lags it reduces to ordinary averaging. It sends one extra number with each model.

\paragraph{A two-client example.} Clients A and B start with lag zero. A is unavailable for 8 rounds, so $\ell_A=8$, while B keeps training and stays at zero. In the round in which A returns, each averages the two models. Ordinary averaging gives A's own model weight one half in A's average and in B's; both end with lag 4, although B missed nothing. \sfgl gives A's model weight $0.5^{8}/(1+0.5^{8})\approx0.4$\% in both averages. A largely adopts B's model, B is almost unaffected, and both end with lag 0.03.

\paragraph{Availability schedules.} \emph{FLASH replay} replays a recorded smartphone availability trace \citep{yang2021flash}. Each simulated client is assigned one phone and is available in a round when that phone was charging, on Wi-Fi and had its screen off; one round represents ten minutes of the trace. The trace decides only which simulated clients take part in which round. All models were trained on server GPUs, and nothing here measures on-device training or hardware failures. The \emph{synthetic} schedule draws each client's availability from an independent two-state Markov chain. In the 100-client setting the two schedules have similar shares of rounds offline, 82\% and 80\%, but mean offline spells of 46.2 and 9.6 rounds; they also differ in how often clients are available together, how unevenly clients take part, and participation near the end of a run (\cref{app:traces}). Three further schedules derived from these are introduced in \cref{sec:qwen-schedule}.

\paragraph{Unit of analysis.} A seed fixes the data split, the graph, the availability traces and the initial model. The two rules of a comparison share the seed, and we analyse paired differences with seeds as the unit; intervals are 95\% $t$-intervals over seeds unless stated. Differences in accuracy are in percentage points. Differences in perplexity are percentage reductions, $100\,(1-\mathrm{ppl}_{\text{\sfgl}}/\mathrm{ppl}_{\text{\dfedavg}})$. Learning rates were selected for each rule on validation data; every comparison below states which rates it uses.

\section{Fine-tuning Qwen2.5-7B across clients with different task mixtures}\label{sec:qwen}

\begin{figure*}[t]
\centering
\includegraphics[width=\textwidth]{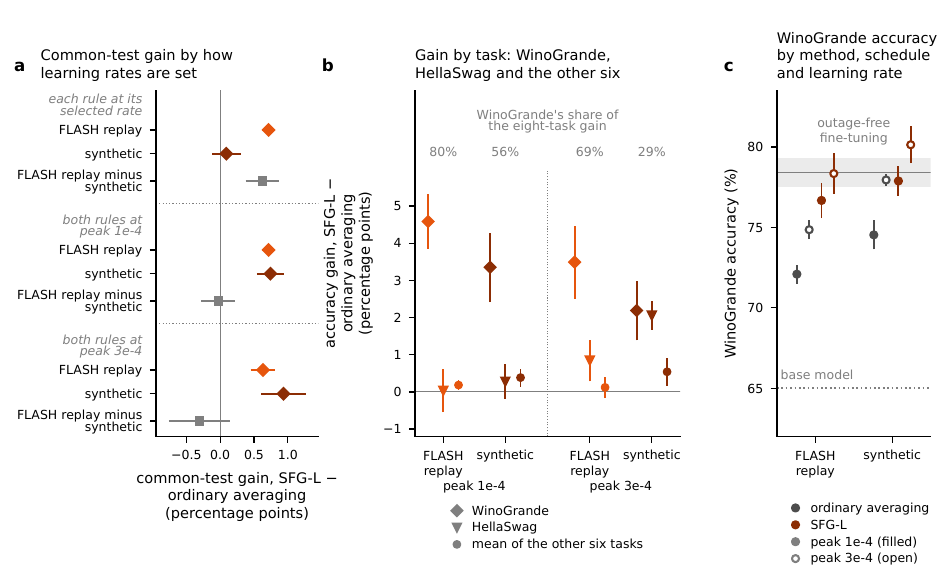}
\caption{\textbf{Qwen2.5-7B fine-tuning.} (a)~\sfgl minus ordinary averaging on the task-balanced common test under each schedule (diamonds), and the paired difference in gain between schedules (squares). Validation selected $10^{-4}$ for both rules under FLASH replay. Under synthetic outages it selected $3\times10^{-4}$ for ordinary averaging and $10^{-4}$ for \sfgl. The next two groups match the rates across rules. (b)~Gains on WinoGrande, HellaSwag and the mean of the six remaining tasks for all four matched comparisons. Percentages give WinoGrande's share of the eight-task gain. (c)~WinoGrande accuracy for the base model, outage-free fine-tuning at $10^{-4}$, and each rule under outages. Error bars are 95\% intervals over six seeds.}
\label{fig:qwen}
\end{figure*}

\subsection{Experiment and outcomes}\label{sec:qwen-setup}
100 clients on a 6-nearest-neighbour graph adapt Qwen2.5-7B \citep{qwen2024qwen25} with LoRA adapters of rank 8 \citep{hu2022lora}. Each client holds its own mixture of eight multiple-choice tasks, with proportions drawn from a Dirichlet distribution ($\alpha=0.3$): BoolQ \citep{clark2019boolq}, PIQA \citep{bisk2020piqa}, Social IQa \citep{sap2019siqa}, HellaSwag \citep{zellers2019hellaswag}, WinoGrande \citep{sakaguchi2020winogrande}, ARC-Easy and ARC-Challenge \citep{clark2018arc} and OpenBookQA \citep{mihaylov2018obqa}. A run is 300 rounds of 10 AdamW steps. Clients exchange adapters only; the two LoRA factors are averaged separately, and both rules average the same adapter representation, so they differ only in the weights.

The \emph{primary outcome} is the typical final client's accuracy on a task-balanced test: each of the 100 final client adapters is evaluated on the same 4,000 held-out items, 500 from each task, and the client scores are averaged. We call this \emph{common-test accuracy}. Two secondary outcomes answer other deployment questions: each client's accuracy on test items from its \emph{own task mixture}, and the common-test accuracy of the single \emph{averaged adapter} obtained by averaging all clients' adapters. As a check on general language modelling we also measure next-token loss on WikiText-103 \citep{merity2017wikitext}.

Each rule's learning-rate schedule was selected on validation data, for each availability schedule, from cosine schedules with peak rates $10^{-4}$ and $3\times10^{-4}$. Under FLASH replay both rules selected $10^{-4}$. Under the synthetic schedule \sfgl selected $10^{-4}$ and ordinary averaging $3\times10^{-4}$. Outage-free fine-tuning, the reference, uses ordinary averaging at $10^{-4}$. Because the selected rates differ, we also ran both rules at both rates under both schedules, with six seeds each, and state the rates of every comparison. \Cref{sec:qwen-avg,sec:qwen-tasks,sec:qwen-other} compare the rules at matched rates, mainly at $10^{-4}$; \cref{sec:qwen-lr} turns to the learning rate itself.

\subsection{The average effect at matched learning rates}\label{sec:qwen-avg}
\paragraph{FLASH replay.} With both rules at $10^{-4}$, \sfgl improved common-test accuracy by 0.72 percentage points over six seeds (0.62 to 0.81; \cref{fig:qwen}a). The first three seeds gave 0.70. That result motivated a repeat on three additional seeds with the schedules fixed beforehand, which gave 0.73 (0.49 to 0.97; \cref{fig:schedules}a). Outages cost ordinary averaging 1.22 points relative to outage-free fine-tuning, so the rule recovered about three fifths of that cost. With both rules at $3\times10^{-4}$ the difference was 0.63 points (0.45 to 0.82).

\paragraph{Synthetic schedule.} With both rules at $10^{-4}$ the difference was 0.74 points (0.54 to 0.94; six seeds), and with both at $3\times10^{-4}$ it was 0.94 (0.61 to 1.26). The gain is thus present at both rates under both schedules, and we detected no difference between the schedules, although the intervals allow one in either direction: the paired difference between FLASH replay and the synthetic schedule is $-$0.03 points ($-$0.28 to 0.23) at $10^{-4}$ and $-$0.30 ($-$0.76 to 0.15) at $3\times10^{-4}$ (\cref{fig:qwen}a). At $10^{-4}$ outages cost ordinary averaging 0.50 points (0.18 to 0.82) under the synthetic schedule.

\subsection{Which tasks account for the gain}\label{sec:qwen-tasks}
We examined the original test items after seeing the aggregate result. We then specified a test of the WinoGrande gain under FLASH replay on additional items before scoring them. Both analyses used saved outputs or adapters.

\paragraph{Four accuracy levels on WinoGrande.} Under FLASH replay WinoGrande gained 4.58 points (3.85 to 5.32) and BoolQ 1.45 (1.02 to 1.89); only these two task differences survived Holm correction across the eight tasks (\cref{tab:qwen-shared-tasks}; see also \cref{fig:tasks-rates}a). The levels behind the WinoGrande difference are the most informative numbers we have (\cref{fig:qwen}c). The base model scores 65.0\%. Outage-free fine-tuning reaches 78.4\%. Under FLASH replay ordinary averaging reaches 72.1\% and \sfgl 76.7\%. Relative to the 13.4-point improvement of the outage-free reference, the two rules retain about 53\% and 87\%. These percentages describe endpoint accuracy on this task, with the base model as the origin.

\paragraph{Breadth across clients, concentration across tasks.} About 96\% of clients improved on WinoGrande, and the median client improved on WinoGrande by 4.7 points. Because the common test weights tasks equally, WinoGrande contributes about 80\% of the eight-task gain. Two task-removal analyses show how a reader's conclusion would change with the composition of the benchmark. Across the other seven tasks the gain is 0.16 points (0.08 to 0.25). Removing BoolQ as well leaves $-0.05$ ($-0.14$ to 0.04) across the remaining six.

\paragraph{The other matched comparisons.} The pattern repeats in two of the three other matched comparisons. Under the synthetic schedule at $10^{-4}$ WinoGrande gains 3.35 points (2.42 to 4.29) and BoolQ 1.52; under FLASH replay at $3\times10^{-4}$ they gain 3.49 and 1.27; in both, no other task's interval excludes zero after correction. Under the synthetic schedule at $3\times10^{-4}$ the gain is spread more widely: HellaSwag gains 2.06 points (1.67 to 2.44), about as much as WinoGrande (2.19), the seven tasks other than WinoGrande gain 0.76 points on average, and WinoGrande contributes 29\% of the eight-task gain, against 56\% and 69\% in the other two (\cref{fig:qwen}b; \cref{app:qwen-shared}). How concentrated the gain is therefore depends on the configuration.

\paragraph{What the pattern does and does not explain.} WinoGrande is the task on which fine-tuning adds most, which suggests that tasks with more to learn benefit more. HellaSwag sets a limit to that explanation: outage-free fine-tuning improves it by 4.2 points, yet under FLASH replay at $10^{-4}$ \sfgl improves on ordinary averaging by only 0.03 points ($-0.54$ to 0.61), whereas under the synthetic schedule at the higher rate it gains about two points. Adaptation produced measurable gains on both tasks, and where the advantage of weighting falls depended on the configuration. We therefore treat ``tasks with more to learn benefit more'' as a hypothesis for another task suite.

\paragraph{Clients' own data and prompt format.} Clients with less WinoGrande in their own training mixture tended to gain more on WinoGrande: the gain falls by 0.26 points for each additional tenth of the mixture (0.10 to 0.42). This is consistent with learning through peers, although we have no evidence about which neighbours contributed. The gain appeared with either correct answer position. Re-scoring the saved adapters gave a gain of 4.30 points when the options were reversed and 4.55 when the instruction was reworded (\cref{app:qwen-wino}).

\paragraph{What ``retained'' means at the level of items.} To test whether the WinoGrande difference is retention of what fine-tuning corrects, we grouped each client's 500 WinoGrande items by whether the base model and that client's adapter from outage-free fine-tuning answer them correctly. Outage-free fine-tuning corrected 19.0\% of the items, which the base model missed, and broke 5.5\%, which the base model had answered. On the corrected items, the adapters trained under FLASH replay were right on 51.9\% with ordinary averaging and on 78.4\% with \sfgl. This group contributes $+5.01$ points (4.48 to 5.54) to the 4.58-point difference. The other groups contribute $-0.86$ (items that fine-tuning broke), $-0.56$ (items that both references answer correctly) and $+0.99$ (items that both miss). The difference is thus mainly corrections that \sfgl's adapters keep and ordinary averaging's lose, partly offset by errors of outage-free fine-tuning that they also keep (\cref{app:qwen-items}).

\paragraph{Additional WinoGrande items.} The common test uses 500 of the 1,267 eligible WinoGrande items. To test whether the finding depends on that draw, we evaluated the saved adapters from the 30 runs described in Appendix~\ref{app:qwen-heldout} on the remaining 767 items, with the scoring, the item groups and the reading of the result fixed beforehand, and after confirming that the evaluation code reproduces every stored prediction on the original items. Under FLASH replay the difference on the additional items is 4.23 points (3.23 to 5.23), against 4.58 on the original ones; the two differ by $-0.36$ points ($-0.78$ to 0.07). Some of the additional items are in clients' own test sets; excluding every such item leaves 4.17 points (2.98 to 5.36). The levels repeat as well: 65.1\% for the base model, 76.8\% for outage-free fine-tuning, 70.9\% for ordinary averaging and 75.1\% for \sfgl, so the two rules retain 49\% and 85\% of the improvement. 96.5\% of clients improve on these items, and the corrected items again carry the difference (\cref{app:qwen-heldout}).

\subsection{Other outcomes and general language modelling}\label{sec:qwen-other}

\begin{table*}[t]
\centering\small
\begin{tabular}{@{}lrrrr@{}}
\toprule
 & \multicolumn{3}{c}{Accuracy (points)} & WikiText-103\\
\cmidrule(lr){2-4}
 & Common test & Own task mixture & Averaged adapter & loss (nats)\\
\midrule
\multicolumn{5}{@{}l}{\emph{Both rules at $10^{-4}$}}\\
\quad FLASH replay & 0.72 {\scriptsize[0.62, 0.81]} & 0.52 {\scriptsize[0.39, 0.65]} & 1.36 {\scriptsize[1.21, 1.51]} & $+$0.082 {\scriptsize[0.063, 0.101]}\\
\quad Synthetic & 0.74 {\scriptsize[0.54, 0.94]} & 0.55 {\scriptsize[0.27, 0.82]} & 0.79 {\scriptsize[0.45, 1.12]} & $+$0.034 {\scriptsize[0.021, 0.047]}\\
\multicolumn{5}{@{}l}{\emph{Both rules at $3\times10^{-4}$}}\\
\quad FLASH replay & 0.63 {\scriptsize[0.45, 0.82]} & 0.24 {\scriptsize[$-$0.18, 0.66]} & 1.46 {\scriptsize[1.22, 1.70]} & $+$0.273 {\scriptsize[0.130, 0.416]}\\
\quad Synthetic & 0.94 {\scriptsize[0.61, 1.26]} & 0.54 {\scriptsize[$-$0.02, 1.10]} & 1.03 {\scriptsize[0.62, 1.44]} & $+$0.036 {\scriptsize[$-$0.037, 0.108]}\\
\multicolumn{5}{@{}l}{\emph{Ordinary averaging at $3\times10^{-4}$, \sfgl at $10^{-4}$}}\\
\quad FLASH replay & 0.22 {\scriptsize[0.09, 0.36]} & 0.02 {\scriptsize[$-$0.17, 0.21]} & 0.71 {\scriptsize[0.41, 1.02]} & $-$0.025 {\scriptsize[$-$0.035, $-$0.016]}\\
\quad Synthetic & 0.09 {\scriptsize[$-$0.12, 0.30]} & 0.00 {\scriptsize[$-$0.37, 0.37]} & 0.07 {\scriptsize[$-$0.19, 0.34]} & $-$0.074 {\scriptsize[$-$0.096, $-$0.051]}\\
\bottomrule
\end{tabular}
\caption{Qwen2.5-7B, six seeds: \sfgl minus ordinary averaging at the stated peak learning rates, with 95\% intervals. \emph{Common test}: client adapters on the task-balanced test, the primary outcome. \emph{Own task mixture}: client adapters on their own test items. \emph{Averaged adapter}: the average of all clients' adapters on the common test. For WikiText-103 loss of the client adapters, higher means worse next-token prediction on that corpus (base model 2.321, outage-free fine-tuning 2.426).}
\label{tab:qwen}
\end{table*}

\paragraph{Own task mixture and averaged adapter.} At $10^{-4}$ \sfgl also improved clients' accuracy on their own task mixtures, by 0.52 points under FLASH replay and 0.55 under the synthetic schedule, and the accuracy of the averaged adapter, by 1.36 and 0.79 points (\cref{tab:qwen}). At $3\times10^{-4}$ the averaged adapter again gained, by 1.46 and 1.03 points; the estimates on clients' own mixtures were positive, with intervals that include zero.

\paragraph{A measured trade-off on WikiText-103.} At matched rates \sfgl ended with higher next-token loss on WikiText-103 than ordinary averaging in all four comparisons. At $10^{-4}$ the differences were 0.082 nats per token (0.063 to 0.101) under FLASH replay and 0.034 (0.021 to 0.047) under the synthetic schedule. At $3\times10^{-4}$ they were 0.273 (0.130 to 0.416) and 0.036 ($-0.037$ to 0.108), respectively. Under FLASH replay at $10^{-4}$ \sfgl finished close to the outage-free reference ($+$0.005 nats from it), whereas ordinary averaging finished 0.077 nats below it. The rule that kept more of the task gain also moved further from the base model's loss on this corpus. Loss on one corpus does not measure general language ability.

\subsection{The learning rate}\label{sec:qwen-lr}
\paragraph{A higher rate helps both rules under outages.} Tripling the peak learning rate raised the common-test accuracy of ordinary averaging by 0.49 points (0.45 to 0.54) under FLASH replay and 0.65 (0.50 to 0.80) under the synthetic schedule, and that of \sfgl by 0.41 (0.23 to 0.59) and 0.85 (0.66 to 1.03). Outage-free fine-tuning did not gain ($-$0.20; $-$0.69 to 0.29). The higher rate also raised WikiText-103 loss in every case: by 0.11 nats for ordinary averaging under both schedules, by 0.30 and 0.11 for \sfgl, and by 1.22 for outage-free fine-tuning. Under outages a higher rate thus buys task accuracy at a cost in loss on general text.

\paragraph{Weighting against a baseline at three times the rate.} For ordinary averaging, a threefold higher learning rate recovers most of what weighting by lag gains at the lower rate. With \sfgl at $10^{-4}$ and ordinary averaging at $3\times10^{-4}$, the difference under FLASH replay was 0.22 points (0.09 to 0.36), positive in all six seeds; this comparison and its reading were fixed before the runs. Under the synthetic schedule it was 0.09 ($-$0.12 to 0.30; \cref{tab:qwen}). In both, \sfgl had the lower WikiText-103 loss, by 0.025 and 0.074 nats (\cref{tab:qwen}).

\paragraph{The selection criterion decides which comparison is reported.} Our criterion, fixed before the experiment, was validation accuracy averaged over training. It selected $10^{-4}$ for both rules under FLASH replay and different rates for the two rules under the synthetic schedule. Had we compared only those configurations, we would have reported 0.72 points under FLASH replay against 0.09 under the synthetic schedule, a post hoc paired difference of 0.62 points (0.38 to 0.87). At either matched rate we detected no difference in the gain between schedules (\cref{fig:qwen}a). Validation accuracy at the end of training would instead have selected $3\times10^{-4}$ for both rules under both schedules. Both rules were more accurate at the higher of the two rates, so this grid does not bracket either rule's best rate, and we do not know how the comparison changes at higher rates (\cref{app:qwen-shared}).

\subsection{Other availability schedules}\label{sec:qwen-schedule}
FLASH replay differs from the synthetic schedule in several respects: offline spells almost five times as long, synchronised availability, more uneven participation and fewer clients available late in a run. To see whether the gain depends on any of these, we ran both rules at $10^{-4}$ on three further schedules with six seeds each (\cref{fig:schedules}a). In the first, each client's FLASH trace is shifted in time by its own random offset, which removes the synchrony between clients. In the second, all traces are shifted by one common offset, which keeps the synchrony and moves the spells within training. The third is a synthetic schedule with independent clients, matched to FLASH replay in the share of rounds offline and the mean offline spell.

The gain is present on every schedule: 0.89, 0.78 and 1.05 points on the three further ones. Three contrasts had been fixed beforehand, each with a margin of 0.25 points for calling two gains equivalent: FLASH replay against the per-client shift ($-0.17$ points, $-0.52$ to 0.17), the per-client shift against the matched synthetic schedule ($-0.16$, $-0.55$ to 0.23), and the common offset against the per-client shift ($-0.11$, $-0.43$ to 0.21). All three were inconclusive: no interval excludes zero, and none is narrow enough to establish equivalence. These experiments do not identify a property of the schedule that the gain depends on; they show that it does not require the recorded trace, its synchrony or long offline spells. On every schedule WinoGrande again gains most (\cref{app:traces}).

\section{Pretraining GPT-124M: duration and checkpoint maturity}\label{sec:gpt}

\begin{figure*}[t]
\centering
\includegraphics[width=\textwidth]{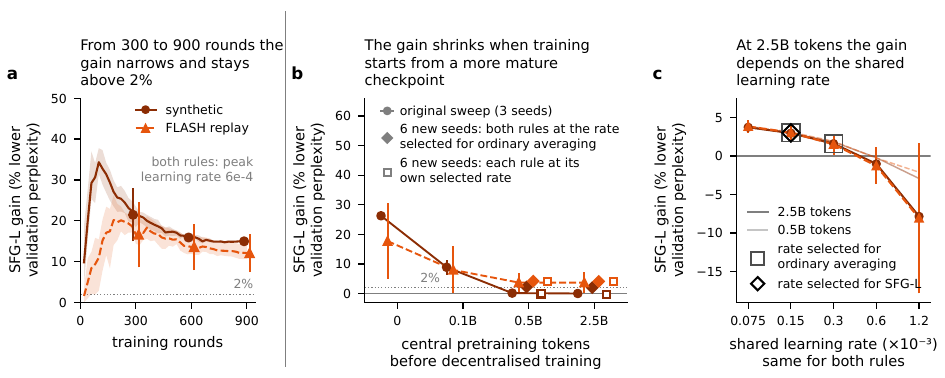}
\caption{\textbf{GPT-124M pretraining: reduction in validation perplexity with \sfgl relative to ordinary averaging.} Three separate experiments. (a)~\emph{Duration}: from random initialisation, both rules at a peak learning rate of $6\times10^{-4}$, over 900 rounds. Lines are means over three seeds and bands the range over seeds; markers give the reduction at 300, 600 and 900 rounds. (b)~\emph{Starting checkpoint}: decentralised training begins from random initialisation or from a centrally pretrained checkpoint. Lines: the original three seeds, each rule at its own selected rate. Diamonds and squares: a repeat on six new seeds, with both rules at the baseline's rate and with each rule at its own. (c)~\emph{Learning rate}: from the 2.5-billion-token checkpoint (lighter lines: 0.5 billion), both rules at the same peak rate, on the three seeds used to select rates. Error bars are 95\% intervals over seeds.}
\label{fig:gpt}
\end{figure*}

The fine-tuning study starts from a mature model. The pretraining study asks two questions that it cannot: whether an advantage persists as training continues, and how it changes with the maturity of the checkpoint from which decentralised training starts. 16 clients on a 3-nearest-neighbour graph train GPT-124M, the GPT-2-small architecture \citep{radford2019gpt2}, on FineWeb-Edu \citep{penedo2024fineweb} with identically distributed data, 50 AdamW steps per round, under the same two availability schedules. Clients exchange full models. We report the percentage by which \sfgl lowers validation perplexity, averaged over clients. These experiments differ from the fine-tuning study in architecture, data, number of clients, objective and what is exchanged, so we compare within GPT only.

\paragraph{Does the advantage persist when training runs longer?} We trained from random initialisation for 900 rounds, at equal tokens, with both rules at a shared peak rate of $6\times10^{-4}$ and a cosine schedule over the whole run. Within this experiment the advantage declined from 19.0\% at 300 rounds to 14.7\% at 600 and 13.5\% at 900 (11.4 to 15.6; \cref{fig:gpt}a). At the end it was positive under both schedules, 15.0\% under the synthetic schedule and 12.1\% under FLASH replay, and all six seed-and-schedule pairs exceeded the 2\% we had set beforehand as the smallest reduction of interest. The advantage therefore persisted through the window we measured; it was still narrowing at the end, and we cannot say whether it would close with longer training. A separate 300-round experiment, with separately selected rates and a schedule that decays within 300 rounds, gave 22.0\%; we use the three values from the long run to describe the trajectory.

\paragraph{What happens from a more mature checkpoint?} We pretrained the same model centrally and started 300-round decentralised training from checkpoints saved after 0.1, 0.5 and 2.5 billion tokens. On the original three seeds, with each rule at its selected rate, the advantage fell from 22.0\% at random initialisation to 8.4\%, 1.9\% and 1.8\% (\cref{fig:gpt}b), and at the last start its interval included zero ($-0.5$ to 4.2). We therefore repeated the two most mature starts on six new seeds, after selecting each rule's rate on three further seeds. From the 2.5-billion-token checkpoint, with both rules at the baseline's selected rate, the advantage was 3.0\% (2.5 to 3.6).

\paragraph{The baseline's learning rate decides part of the answer.} Under the synthetic schedule the two rules selected different rates at the mature checkpoint, $3\times10^{-4}$ for ordinary averaging and $1.5\times10^{-4}$ for \sfgl. There the advantage was 2.0\% with both rules at the baseline's rate and $-0.2$\% ($-0.7$ to 0.2) with each rule at its own. Under FLASH replay both rules selected the same rate and the advantage was 4.1\%. On the seeds used for selection, the advantage at a shared rate fell steadily as that rate rose, from about 3.8\% at $7.5\times10^{-5}$ to a disadvantage of about 8\% at $1.2\times10^{-3}$ (\cref{fig:gpt}c). At a mature checkpoint, a statement about the advantage is therefore also a statement about the learning rate, as in fine-tuning (\cref{sec:qwen-lr}). A vision control makes the same point about the cost of outages itself (\cref{app:vision}).

\paragraph{A reference with missed communication restored.} Ordinary averaging with the model exchanges restored, in which every client averages in every round and only available clients train, reached 21.7\% lower perplexity than ordinary averaging under outages from random initialisation and 1.4\% from the 2.5-billion-token checkpoint. \sfgl under outages was 1.6\% (1.0 to 2.2) below that reference at the mature checkpoint, so the reference is not a ceiling (\cref{app:gpt-split}).

\section{Discussion and conclusion}\label{sec:discussion}

\paragraph{What was learned about this benchmark.} At matched learning rates, the improvement in average fine-tuning accuracy repeated on additional training seeds, at a second learning rate and under every availability schedule we tried. A task-level analysis showed that freshness-weighted averaging preserved more of the WinoGrande accuracy gain achieved by outage-free fine-tuning, and an item-level analysis that the difference consists mainly of answers which that fine-tuning corrects. The WinoGrande result repeated on additional items of the same task and under two other prompt formats. This advantage appeared across clients, while the mean effect on the other seven tasks was small in three of the four matched comparisons. An eight-task average of 0.72 points and a seven-task average of 0.16 points describe the same adapters; which number a reader takes away depends on the composition of the benchmark. The rule that kept more of the task gain also ended with higher loss on WikiText-103.

\paragraph{What the learning rate changes.} A threefold higher learning rate improved both rules under outages, by about as much as weighting did, and raised loss on WikiText-103. Our validation criterion gave the two rules the same rate under FLASH replay and different rates under the synthetic schedule. Comparing only those configurations would have suggested that the rule helps under recorded smartphone availability and not under synthetic outages; at matched rates it helps under both. We therefore claim an advantage at matched learning rates, on every schedule tested. Against a baseline given three times the rule's learning rate, the advantage was 0.22 points under FLASH replay and not distinguishable from zero under the synthetic schedule, with lower loss on WikiText-103 for \sfgl in both. Both rules were best at the highest rate we tried, so we make no claim about rates beyond it.

\paragraph{Pretraining.} In separate GPT pretraining experiments, a perplexity advantage persisted through 900 rounds and depended on the learning-rate choice at more mature starting checkpoints.

\paragraph{Implications.} These findings support evaluating participation-aware aggregation against a tuned baseline, at matched learning rates as well as at separately selected ones, and reporting which tasks account for its aggregate effect. A difference between availability conditions should be read as an effect of availability only when the learning rates are the same in both.

\section*{Limitations}\label{sec:limitations}

\paragraph{Simulation.} All experiments are simulations on server GPUs. The FLASH trace is a recording of when phones were charging, on Wi-Fi and idle; it determines which simulated clients take part in a round. On-device runtime, energy use, network cost and hardware failures were outside the scope of the simulations.

\paragraph{Learning rates.} We compared two peak learning rates in fine-tuning. Under outages both rules were more accurate at the higher one, so the grid does not bracket either rule's best rate, and the comparison could change at higher rates; in pretraining from a mature checkpoint the advantage at a shared rate shrank as the rate grew and reversed at the highest. The rates that enter a comparison at separately selected rates depend on the selection criterion: ours, fixed beforehand, chose different configurations from those that final validation accuracy would have chosen. The task breakdown covers both rates. The WinoGrande follow-ups used the validation-selected rates for each schedule. The primary test on additional items used $10^{-4}$ under FLASH replay. In pretraining, the comparison from random initialisation over 300 rounds and that from mature checkpoints under the synthetic schedule use separately selected rates and mix the effect of the rule with that of the rate; we report shared-rate comparisons where we have them.

\paragraph{One benchmark for fine-tuning.} The fine-tuning findings rest on Qwen2.5 models, LoRA of rank 8 and eight English multiple-choice tasks, of which one carried most of the gain in three of four matched comparisons. Another task mixture could give a different average, and we did not test generation tasks or other languages. The task-level analyses followed inspection of the aggregate result. The WinoGrande finding was then confirmed on additional items of that task with the analysis fixed beforehand, which shows that it does not depend on the items drawn; whether it extends to other tasks remains to be tested on another task suite.

\paragraph{Availability schedules.} The gain at a shared rate of $10^{-4}$ was present on five schedules, but the contrasts designed to isolate synchrony and the structure of the recorded trace were inconclusive, so we cannot say which properties of a schedule, if any, change its size. All five schedules have about four fifths of rounds offline; we did not vary that share in fine-tuning.

\paragraph{Pretraining.} Maturity was varied in one architecture of 124M parameters with identically distributed data and 16 clients. The long runs cover 900 rounds, over which the advantage narrowed. The pretraining and fine-tuning studies differ in too many respects to attribute the difference between their results to pretraining versus fine-tuning.

\paragraph{General language modelling.} We measured next-token loss on one corpus. It does not measure general language ability.

\section*{Acknowledgements}
This work was supported by the European Union's Horizon Europe research and innovation programme through the ELLIE project (grant agreement No.~101178099) and the TERRA project (grant agreement No.~101189962). Views and opinions expressed are however those of the author(s) only and do not necessarily reflect those of the European Union or the granting authority. Neither the European Union nor the granting authority can be held responsible for them.

\appendix

\section{Where to find the details}\label{app:overview}
The appendix follows the order of the main text. \Cref{app:qwen} covers Qwen2.5-7B fine-tuning: model, adapters, tasks, scoring, the construction of the common test and the selection of learning rates (\cref{app:qwen-setup}); seed-level estimates, the pooled estimate and the overlap sensitivity (\cref{app:qwen-seeds}); absolute scores and gains for every task (\cref{app:qwen-tasks}); the checks on WinoGrande (\cref{app:qwen-wino}); the item-level analysis (\cref{app:qwen-items}); WinoGrande on additional items (\cref{app:qwen-heldout}); WikiText-103 (\cref{app:qwen-retention}); the full grid of learning rates (\cref{app:qwen-shared}); the smaller model (\cref{app:qwen-small}); and the availability schedules, including the three further ones (\cref{app:traces}). \Cref{app:gpt} covers GPT-124M pretraining: setup and the repeated trace (\cref{app:gpt-setup}), the 900-round runs (\cref{app:gpt-long}), the starting-checkpoint experiment and its repeat (\cref{app:gpt-maturity,app:gpt-repeat}), the reference with communication restored (\cref{app:gpt-split}) and the share of rounds offline (\cref{app:gpt-offline}). \Cref{app:vision} gives the one vision control used in the text and \cref{app:repro} the materials and compute.

\paragraph{Evidential status of the results.} The following were tested with the design and the reading of the result fixed beforehand: the Qwen gain under FLASH replay on three additional seeds (schedules fixed); the signs of the WikiText-103 difference in the two selected configurations (the same additional seeds); the WinoGrande difference on 767 further items; the contrast between FLASH replay and the synthetic schedule at the shared peak rate of $10^{-4}$; the comparison of \sfgl with ordinary averaging at three times its learning rate under FLASH replay; the three contrasts between availability schedules, which were inconclusive; the GPT advantage from mature checkpoints (six new seeds, rates selected on three others); and the 900-round GPT runs, which reuse the seeds of the 300-round experiment with a threshold set beforehand. The task breakdown, the first checks on WinoGrande, the item-level analysis, the comparison of selection criteria, the check under other prompt formats, the between-schedule contrasts at the selected rates and at the higher matched rate, and the advantage at each shared learning rate in GPT followed inspection of earlier results and are descriptive.

\section{\texorpdfstring{Qwen2.5-7B fine-tuning: details for \cref{sec:qwen}}{Qwen2.5-7B fine-tuning details}}\label{app:qwen}

\subsection{Setup}\label{app:qwen-setup}
\paragraph{Model and adapters.} The model is \texttt{Qwen/Qwen2.5-7B}, the base model and not the instruction-tuned variant \citep{qwen2024qwen25}, at Hugging Face snapshot {\small\texttt{d149729398750b98c0af\allowbreak 14eb82c78cfe92750796}}, in bfloat16 with the backbone frozen and its own tokenizer; the adapter parameters are kept in 32-bit floating point. LoRA \citep{hu2022lora} has rank 8 and $\alpha=16$ on the query, key, value and output projections of all 28 layers. Clients exchange adapters only. The two LoRA factors are averaged separately, for both rules. Local training uses AdamW \citep{loshchilov2019adamw} with no weight decay, gradient clipping at 1.0 and 10 steps per round, each on 4 items drawn with replacement from the client's training items; the optimiser state is reset every round. A run is 300 rounds. 100 clients sit on a 6-nearest-neighbour geometric graph.

\paragraph{Tasks and sources.} \Cref{tab:sources} lists the eight datasets as read from the Hugging Face Hub. The training split of each dataset feeds the clients' training and validation data. Test items come from the public validation split for BoolQ, PIQA, Social IQa, HellaSwag and WinoGrande, and from the labelled test split for ARC-Easy, ARC-Challenge and OpenBookQA; below we call this the evaluation split. The dataset revisions were not written down when the data were prepared. Re-running the preparation from the single cached snapshot of each dataset reproduces the stored files exactly, and \cref{tab:sources} gives those snapshots.

\begin{table*}[t]
\centering\footnotesize\setlength{\tabcolsep}{4pt}
\begin{tabular}{@{}lllrlrr@{}}
\toprule
Task & Hub dataset & Commit & Train.\ pool & Eval.\ split & Eval.\ pool & Options\\
\midrule
BoolQ & \texttt{google/boolq} & \texttt{35b264d0} & 9,427 & validation & 3,270 & 2\\
PIQA & \texttt{ybisk/piqa} & \texttt{142c5123} & 16,113 & validation & 1,838 & 2\\
Social IQa & \texttt{allenai/social\_i\_qa} & \texttt{537a2ec8} & 33,410 & validation & 1,954 & 3\\
HellaSwag & \texttt{Rowan/hellaswag} & \texttt{218ec52e} & 39,905 & validation & 10,042 & 4\\
WinoGrande & \texttt{allenai/winogrande (xl)} & \texttt{01e74176} & 40,398 & validation & 1,267 & 2\\
ARC-Easy & \texttt{allenai/ai2\_arc} & \texttt{210d026f} & 2,251 & test & 2,376 & 3--5\\
ARC-Challenge & \texttt{allenai/ai2\_arc} & \texttt{210d026f} & 1,119 & test & 1,172 & 3--5\\
OpenBookQA & \texttt{allenai/openbookqa (main)} & \texttt{388097ea} & 4,957 & test & 500 & 4\\
\bottomrule
\end{tabular}
\caption{Sources of the eight tasks: Hugging Face Hub dataset, first eight characters of the snapshot commit, size of the training split (the pool for clients' training and validation items), which split serves as the evaluation split, its size, and the number of answer options.}
\label{tab:sources}
\end{table*}

\paragraph{Prompt format.} Every task is cast as multiple choice with lettered options. A prompt is a task-specific question text, one line per option of the form ``A.\ \emph{option}'', and a final line ``Answer:''. The question text is the passage and question for BoolQ (options \emph{no}, \emph{yes}); ``Goal:'' and the goal for PIQA; the context and question for Social IQa; the activity and context followed by ``Which ending is the most plausible?'' for HellaSwag; ``Fill in the blank:'' and the sentence for WinoGrande; and ``Question:'' and the question for the ARC tasks and OpenBookQA. Options keep their order in the source. No special tokens are added. A WinoGrande item of the common test reads, verbatim from the source:
\begin{quote}\small\ttfamily
Fill in the blank: They were worried the wine would ruin the bed and the blanket, but the \_ was't ruined.\\
A. blanket\\
B. bed\\
Answer:
\end{quote}
Its correct continuation is `` B''. In training, the loss is taken on the token of the correct letter only.

\paragraph{Scoring.} The model reads the prompt, truncated on the left to its last 255 tokens where necessary, and we take the log-probabilities of the option letters as the next token after ``Answer:''. Each letter is a single token. The prediction is the option with the highest log-probability, with ties going to the first option. The same rule scores validation items, clients' own test items and the common test, for client adapters and for the averaged adapter. Left truncation affects 22 BoolQ, 4 PIQA and 31 HellaSwag items of the common test and no WinoGrande item. Mean chance accuracy on the common test is 35.4\%.

\paragraph{Clients' data.} Each client draws a mixture over the eight tasks from a Dirichlet distribution with $\alpha=0.3$ and makes 600 draws from the tasks' training pools according to it: 510 training and 90 validation items. The first 15\% of each task's shuffled training pool is reserved for validation, and training items with the same content as a reserved item are removed, so no client trains on any validation item. Each client also has 50 test items drawn without replacement from the tasks' evaluation splits by its own mixture; these give the \emph{own task mixture} outcome. Different clients' test sets may overlap.

\paragraph{The common test.} For each task we take the evaluation-split items whose content (prompt and all options) does not occur in the training split, that are the first occurrence of their content, and that can be scored by option letter. The smallest such pool, OpenBookQA's, has 500 items, so we draw 500 items per task without replacement with a fixed seed, giving 4,000 items (\cref{tab:benchmarks}). The set does not depend on the seed, rule or schedule of a run, and no client of any seed trained or validated on any of its items. Clients' own test items come from the same evaluation splits, so the two outcomes overlap: on average about 17 of a client's 50 own test items are also common-test items. \Cref{app:qwen-seeds} gives the common-test result without those items.

\begin{table}[h]
\centering\small\setlength{\tabcolsep}{2.5pt}
\begin{tabular}{@{}lrrrr@{}}
\toprule
Task & Eval.\ & Eligible & Common & Base\\
 & split & items & test & acc.\ (\%)\\
\midrule
BoolQ & 3,270 & 3,270 & 500 & 86.8\\
PIQA & 1,838 & 1,837 & 500 & 85.4\\
Social IQa & 1,954 & 1,930 & 500 & 77.4\\
HellaSwag & 10,042 & 10,042 & 500 & 86.6\\
WinoGrande & 1,267 & 1,267 & 500 & 65.0\\
ARC-Easy & 2,376 & 2,374 & 500 & 96.0\\
ARC-Challenge & 1,172 & 1,172 & 500 & 88.2\\
OpenBookQA & 500 & 500 & 500 & 89.6\\
\bottomrule
\end{tabular}
\caption{The eight tasks: size of the evaluation split used for testing, items eligible for the common test, items drawn, and the base model's accuracy on them before any fine-tuning.}
\label{tab:benchmarks}
\end{table}

\paragraph{Learning rates.} The candidates were cosine schedules with peak rates $3\times10^{-4}$ and $10^{-4}$, both ending at a tenth of the peak. For each rule and availability schedule, and for outage-free fine-tuning, we selected the candidate with the higher validation accuracy averaged over the 30 evaluations of a run (every 10 rounds) and over the first three seeds; ties went to the lower rate. Ordinary averaging under the synthetic schedule selected $3\times10^{-4}$. Every other combination, including outage-free fine-tuning, selected $10^{-4}$. The selections were kept for the three additional seeds.

\paragraph{Seeds.} A seed fixes the task mixtures, the data, the graph and the availability traces, and the two rules share it. The runs of the first three seeds were made for an earlier comparison on clients' own task mixtures and were repeated, with identical results, to add the common-test evaluation. Seeds 4--6 are new runs made after the three-seed result was known.

\subsection{Seed-level estimates and sensitivity analyses}\label{app:qwen-seeds}
\Cref{tab:qwen-seeds} lists the gain for every seed in the two selected configurations (both rules at $10^{-4}$ under FLASH replay; ordinary averaging at $3\times10^{-4}$ and \sfgl at $10^{-4}$ under the synthetic schedule) on the three accuracy outcomes of \cref{tab:qwen}. Under FLASH replay, at a shared learning rate, the common-test gain lies between 0.61 and 0.84 points in all six seeds.

\paragraph{The pooled estimate.} Pooled over the two selected configurations the gain is 0.40 points. We report two 97.5\% intervals and how each was calculated. The interval specified before the three additional seeds were run treats the 12 seed-and-schedule pairs as the units: 0.13 to 0.67. Because the two pairs of a seed share data, graph and initial model, we also average the two schedules within each seed and treat the six seeds as the units: 0.27 to 0.53. Both intervals include the 0.5-point threshold of practical interest, which, like the 97.5\% level, was fixed beforehand. The prediction that the gain would lie below that threshold was therefore inconclusive under either calculation.

\paragraph{Overlap with clients' own test items.} On the common-test items that are in no client's own test set, the pooled gain is 0.51 points (0.22 to 0.79; 97.5\%, pairs as units); \cref{tab:qwen-seeds} lists it per seed in the column ``unseen''.

\begin{table}[h]
\centering\small\setlength{\tabcolsep}{2.5pt}
\begin{tabular}{@{}llrrrr@{}}
\toprule
 & & \multicolumn{2}{c}{Common test} & Own & Averaged\\
Schedule & Seed & all items & unseen & tasks & adapter\\
\midrule
FLASH replay & 1 & $+$0.61 & $+$0.65 & $+$0.44 & $+$1.50\\
 & 2 & $+$0.69 & $+$0.78 & $+$0.42 & $+$1.40\\
 & 3 & $+$0.80 & $+$1.06 & $+$0.58 & $+$1.40\\
 & 4$^{*}$ & $+$0.66 & $+$0.66 & $+$0.72 & $+$1.15\\
 & 5$^{*}$ & $+$0.84 & $+$0.72 & $+$0.58 & $+$1.50\\
 & 6$^{*}$ & $+$0.69 & $+$0.89 & $+$0.40 & $+$1.23\\
\midrule
synthetic & 1 & $+$0.31 & $+$0.56 & $-$0.20 & $+$0.28\\
 & 2 & $-$0.23 & $-$0.18 & $-$0.14 & $-$0.28\\
 & 3 & $-$0.02 & $+$0.07 & $-$0.44 & $-$0.08\\
 & 4$^{*}$ & $+$0.05 & $+$0.10 & $-$0.08 & $+$0.05\\
 & 5$^{*}$ & $+$0.18 & $+$0.62 & $+$0.36 & $+$0.45\\
 & 6$^{*}$ & $+$0.27 & $+$0.14 & $+$0.48 & $+$0.03\\
\bottomrule
\end{tabular}
\caption{Qwen2.5-7B, selected configurations: \sfgl minus ordinary averaging per seed, in percentage points. Under the synthetic schedule ordinary averaging uses a peak rate of $3\times10^{-4}$. \emph{Unseen}: common-test items in no client's own test set. $^{*}$The three additional seeds.}
\label{tab:qwen-seeds}
\end{table}

\subsection{Task breakdown and absolute scores}\label{app:qwen-tasks}
\Cref{tab:qwen-tasks} gives the accuracy of every condition on every task and the gain per task, for the selected configurations; \cref{tab:qwen-shared-tasks} gives the gains for every combination of rates. For the eight per-task tests within an availability schedule we apply Holm's correction \citep{holm1979sequentially}. The base model is already above 80\% on six of the eight tasks, and outage-free fine-tuning adds 3.0 points to the eight-task mean, of which WinoGrande supplies most. Under the synthetic schedule ordinary averaging's eight-task mean is at the level of outage-free fine-tuning, so there is no cost of outages for any rule to recover in that configuration; as \cref{sec:qwen-avg} notes, ordinary averaging ran there at three times the learning rate of the outage-free run.

\begin{table*}[t]
\centering\footnotesize\setlength{\tabcolsep}{3pt}
\begin{tabular}{@{}lrr rrr rrr@{}}
\toprule
 & & & \multicolumn{3}{c}{FLASH replay} & \multicolumn{3}{c}{Synthetic}\\
\cmidrule(lr){4-6}\cmidrule(l){7-9}
Task & Base model & Outage-free & Ord.\ avg. & \sfgl & Gain & Ord.\ avg. & \sfgl & Gain\\
\midrule
BoolQ & 86.8 & 89.7 & 87.1 & 88.6 & $+$1.45 {\scriptsize[1.02, 1.89]}$^{\dagger}$ & 88.9 & 89.0 & $+$0.07 {\scriptsize[$-$0.58, 0.73]}\\
PIQA & 85.4 & 87.6 & 87.1 & 87.4 & $+$0.35 {\scriptsize[$-$0.06, 0.76]} & 88.5 & 88.1 & $-$0.34 {\scriptsize[$-$0.86, 0.17]}\\
Social IQa & 77.4 & 77.7 & 78.4 & 77.9 & $-$0.45 {\scriptsize[$-$0.87, $-$0.02]} & 79.3 & 79.0 & $-$0.22 {\scriptsize[$-$1.05, 0.61]}\\
HellaSwag & 86.6 & 90.8 & 89.1 & 89.2 & $+$0.03 {\scriptsize[$-$0.54, 0.61]} & 89.4 & 90.1 & $+$0.70 {\scriptsize[0.28, 1.11]}\\
WinoGrande & 65.0 & 78.4 & 72.1 & 76.7 & $+$4.58 {\scriptsize[3.85, 5.32]}$^{\dagger}$ & 78.0 & 77.9 & $-$0.06 {\scriptsize[$-$0.74, 0.61]}\\
ARC-Easy & 96.0 & 96.4 & 95.9 & 95.8 & $-$0.05 {\scriptsize[$-$0.30, 0.20]} & 95.8 & 96.0 & $+$0.12 {\scriptsize[$-$0.12, 0.37]}\\
ARC-Challenge & 88.2 & 86.8 & 88.0 & 87.9 & $-$0.16 {\scriptsize[$-$0.44, 0.13]} & 88.2 & 88.4 & $+$0.20 {\scriptsize[$-$0.10, 0.50]}\\
OpenBookQA & 89.6 & 91.4 & 91.3 & 91.3 & $-$0.05 {\scriptsize[$-$0.39, 0.30]} & 91.9 & 92.2 & $+$0.26 {\scriptsize[$-$0.03, 0.56]}\\
\midrule
Eight-task mean & 84.4 & 87.3 & 86.1 & 86.8 & $+$0.72 {\scriptsize[0.62, 0.81]} & 87.5 & 87.6 & $+$0.09 {\scriptsize[$-$0.12, 0.30]}\\
Without WinoGrande & 87.1 & 88.6 & 88.1 & 88.3 & $+$0.16 {\scriptsize[0.08, 0.25]} & 88.9 & 89.0 & $+$0.11 {\scriptsize[$-$0.08, 0.31]}\\
\bottomrule
\end{tabular}
\caption{Qwen2.5-7B, selected configurations (under the synthetic schedule ordinary averaging at a peak rate of $3\times10^{-4}$): common-test accuracy (\%) by task, averaged over clients and six seeds, and the gain of \sfgl in percentage points with 95\% intervals over seeds. \emph{Outage-free}: ordinary averaging with every client available in every round. $^{\dagger}$The interval excludes zero after Holm correction over the eight tasks.}
\label{tab:qwen-tasks}
\end{table*}

\subsection{Checks on the WinoGrande gain}\label{app:qwen-wino}
These analyses were planned and written down after the six-seed result and before any of their quantities was computed; they use the stored per-item outputs of all runs.

\paragraph{Answer position.} Under FLASH replay the gain is 3.75 points (2.10 to 5.39) on items whose correct answer is the first option and 5.42 (4.48 to 6.36) on the others. The share of items on which the first option is chosen differs by $-0.84$ points ($-1.95$ to 0.28).

\paragraph{Prompt format.} We re-scored the saved adapters of the FLASH replay and outage-free runs on the 500 common-test WinoGrande items under two other formats, fixed before the evaluation: the same prompt with the two options in reversed order, and an alternative wording in which the instruction reads ``Choose the option that best fills the blank (\_) in the sentence.'' followed by ``Sentence:'' and the sentence, with the options in the original order. The adapters were trained on the original format only. \sfgl minus ordinary averaging is 4.30 points (3.09 to 5.52) with reversed options and 4.55 (3.78 to 5.32) with the alternative wording, against 4.58 in the original format; the paired changes are $-0.28$ ($-0.97$ to 0.41) and $-0.04$ ($-0.47$ to 0.39). Accuracy itself is lower under the unfamiliar formats, for example 75.0\% and 74.3\% for \sfgl against 76.7\%, but it falls by a similar amount for both rules.

\paragraph{Spread across clients.} 95.7\% of clients have a positive WinoGrande gain (91.7 to 99.6), the median client gains 4.68 points, and the tenth of clients with the largest gains account for 19\% of the total.

\begin{figure}[h]
\centering
\includegraphics[width=\linewidth]{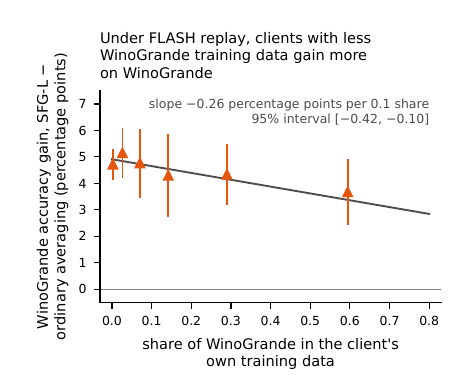}
\caption{Qwen2.5-7B under FLASH replay: clients' WinoGrande gain from \sfgl against the share of WinoGrande in their own training data. Markers are means over the clients in six bins of that share, with 95\% intervals over seeds; the line is the fitted slope.}
\label{fig:winoshare}
\end{figure}

\paragraph{Clients' own WinoGrande data.} \Cref{fig:winoshare} plots clients' WinoGrande gain against the share of WinoGrande in their own training mixture. The slope is $-0.26$ points per tenth of the mixture ($-0.42$ to $-0.10$). Against the share of rounds a client is online the slope is $-0.20$ points per tenth ($-0.44$ to 0.04), which shows no clear relation.

\paragraph{Levels.} Relative to outage-free fine-tuning, ordinary averaging under FLASH replay loses 6.3 points of WinoGrande accuracy and \sfgl 1.7. BoolQ shows the same pattern at a smaller scale (\cref{tab:qwen-tasks}).

\subsection{Item-level analysis: which answers differ}\label{app:qwen-items}
This analysis was planned after the aggregate and task-level results were known and written down before it was computed. For each seed and client, the 500 common-test items of a task are grouped by two references: whether the base model answers the item correctly, and whether the same client's final adapter in the outage-free run of that seed does. Within each group we compare the client's final adapters under ordinary averaging and under \sfgl. A group's contribution to the task difference is its share of the items times the difference in accuracy within it; the four contributions add to the task difference exactly. Client values are averaged within a seed, and intervals are over the six seeds.

\begin{figure}[h]
\centering
\includegraphics[width=\linewidth]{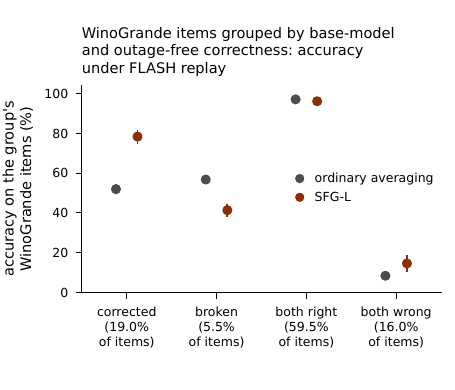}
\caption{Qwen2.5-7B under FLASH replay: accuracy on the four groups of WinoGrande items of \cref{tab:qwen-items}. Error bars are 95\% intervals over seeds.}
\label{fig:items}
\end{figure}

\Cref{tab:qwen-items} gives the result under FLASH replay for WinoGrande and, for comparison, BoolQ and HellaSwag (\cref{fig:items} shows WinoGrande). On all three tasks \sfgl's adapters keep more of the corrections made by outage-free fine-tuning and also more of the errors it introduces. The net difference is large on WinoGrande, where corrections are 19.0\% of the items, and small on HellaSwag, where they are 6.7\% and are offset by the items that fine-tuning broke and by lower accuracy on the items that both references answer correctly. Defining the reference by the majority vote of the 100 outage-free adapters of a seed, so that the grouping is the same for every client, changes little: the corrected group then contributes 4.99 points on WinoGrande. In the synthetic configuration, ordinary averaging already keeps 81.7\% of the WinoGrande corrections and \sfgl 83.8\%, and the corrected group contributes 0.42 points ($-0.33$ to 1.17).

\begin{table*}[t]
\centering\small
\begin{tabular}{@{}lrrrr@{}}
\toprule
Item group & Share of items (\%) & Ordinary averaging (\%) & \sfgl (\%) & Contribution (points)\\
\midrule
\multicolumn{5}{@{}l}{\emph{WinoGrande}}\\
Corrected (base wrong, reference right) & 19.0 & 51.9 & 78.4 & $+$5.01 {\scriptsize[4.48, 5.54]}\\
Broken (base right, reference wrong) & 5.5 & 56.8 & 41.4 & $-$0.86 {\scriptsize[$-$1.07, $-$0.64]}\\
Both right & 59.5 & 97.1 & 96.2 & $-$0.56 {\scriptsize[$-$1.04, $-$0.07]}\\
Both wrong & 16.0 & 8.3 & 14.5 & $+$0.99 {\scriptsize[0.49, 1.49]}\\
\midrule
\multicolumn{5}{@{}l}{\emph{BoolQ}}\\
Corrected (base wrong, reference right) & 5.6 & 46.8 & 73.3 & $+$1.48 {\scriptsize[1.22, 1.75]}\\
Broken (base right, reference wrong) & 2.7 & 46.9 & 38.6 & $-$0.22 {\scriptsize[$-$0.38, $-$0.07]}\\
Both right & 84.1 & 98.2 & 98.1 & $-$0.14 {\scriptsize[$-$0.46, 0.18]}\\
Both wrong & 7.6 & 7.7 & 11.9 & $+$0.33 {\scriptsize[0.00, 0.66]}\\
\midrule
\multicolumn{5}{@{}l}{\emph{HellaSwag}}\\
Corrected (base wrong, reference right) & 6.7 & 60.3 & 70.6 & $+$0.69 {\scriptsize[0.62, 0.76]}\\
Broken (base right, reference wrong) & 2.6 & 48.2 & 34.9 & $-$0.34 {\scriptsize[$-$0.48, $-$0.20]}\\
Both right & 84.0 & 98.7 & 98.2 & $-$0.42 {\scriptsize[$-$0.75, $-$0.10]}\\
Both wrong & 6.7 & 13.1 & 14.9 & $+$0.11 {\scriptsize[$-$0.26, 0.47]}\\
\bottomrule
\end{tabular}
\caption{Qwen2.5-7B under FLASH replay: common-test items grouped by whether the base model and the client's outage-free adapter answer them correctly; accuracy of the adapters trained under outages within each group; and the group's contribution to the task difference (\sfgl minus ordinary averaging), with 95\% intervals over seeds. Contributions add to the task difference.}
\label{tab:qwen-items}
\end{table*}

\setlength{\abovecaptionskip}{4pt}
\setlength{\intextsep}{8pt plus 2pt minus 2pt}
\setlength{\textfloatsep}{10pt plus 2pt minus 2pt}
\renewcommand{\arraystretch}{0.95}
\setlength{\dblfloatsep}{8pt plus 2pt minus 2pt}
\setlength{\dbltextfloatsep}{10pt plus 2pt minus 2pt}
\subsection{WinoGrande on additional items}\label{app:qwen-heldout}
\paragraph{Design.} Of the 1,267 WinoGrande items eligible for the common test (\cref{tab:benchmarks}), 500 were drawn into it. The other 767 are outside the common test and had entered no common-test analysis; some of them are in clients' own test sets, which we treat below. We evaluated on them the saved final client adapters of all 30 runs (both rules under both availability schedules and outage-free fine-tuning, six seeds) and the base model. Before any of these evaluations we fixed the list of items, the scoring rule (that of \cref{app:qwen-setup}), the item groups (those of \cref{app:qwen-items}) and the reading of the result: the finding holds on additional items if the 95\% interval of the difference under FLASH replay excludes zero. As a check on the evaluation code, we first re-scored the 500 original WinoGrande items for all 30 runs and the base model; every stored prediction was reproduced.

\paragraph{Result.} \Cref{tab:qwen-heldout} sets the additional items beside the original ones. Under FLASH replay the difference is 4.23 points (3.23 to 5.23), inside the interval obtained on the original items, and the paired difference between the two item sets is $-0.36$ points ($-0.78$ to 0.07). The item-group pattern is the same (\cref{tab:qwen-heldout-items}): the corrected items contribute $+4.69$ points, and \sfgl's adapters keep 75.8\% of the corrections against 50.3\% for ordinary averaging.

\paragraph{Overlap with clients' own test sets.} Clients' own test items are drawn from the same evaluation split, so some of the additional items are also in a client's own test set: between 249 and 355 of the 767 items are in at least one client's own test set, depending on the seed, and a client has on average between 3.1 and 4.7 of them. No client trains on its test items. Excluding, for each client, the items in its own test set gives 4.22 points (3.22 to 5.23); restricting to the items in no client's own test set gives 4.17 (2.98 to 5.36).

\paragraph{The synthetic configuration.} On the additional items the difference in the synthetic configuration is 0.91 points ($-0.36$ to 2.18), against $-0.06$ on the original items; its interval includes zero, as before. The two item sets differ here by 0.97 points (0.14 to 1.81), so the estimate for this configuration is less stable across items than that under FLASH replay. This comparison was not part of the reading fixed beforehand.

\begin{table}[h]
\centering\small\setlength{\tabcolsep}{2.5pt}
\begin{tabular}{@{}lrr@{}}
\toprule
 & Original 500 & Further 767\\
\midrule
\multicolumn{3}{@{}l}{\emph{FLASH replay}}\\
Difference (points) & 4.58 {\scriptsize[3.85, 5.32]} & 4.23 {\scriptsize[3.23, 5.23]}\\
Base model (\%) & 65.0 & 65.1\\
Outage-free (\%) & 78.4 & 76.8\\
Ordinary averaging (\%) & 72.1 & 70.9\\
\sfgl (\%) & 76.7 & 75.1\\
Retained (\%)$^{\ast}$ & 53 / 87 & 49 / 85\\
Clients that improve (\%) & 95.7 & 96.5\\
\midrule
\multicolumn{3}{@{}l}{\emph{Synthetic configuration}}\\
Difference (points) & $-$0.06 {\scriptsize[$-$0.74, 0.61]} & 0.91 {\scriptsize[$-$0.36, 2.18]}\\
Ordinary averaging (\%) & 78.0 & 75.2\\
\sfgl (\%) & 77.9 & 76.2\\
\bottomrule
\end{tabular}
\caption{Qwen2.5-7B, WinoGrande: the 500 items of the common test and the 767 eligible items outside it, six seeds, with 95\% intervals for the differences. \emph{Difference}: \sfgl minus ordinary averaging. $^{\ast}$Share of the improvement of outage-free fine-tuning over the base model that ordinary averaging / \sfgl keeps.}
\label{tab:qwen-heldout}
\end{table}

\begin{table*}[t]
\centering\small
\begin{tabular}{@{}lrrrr@{}}
\toprule
Item group & Share of items (\%) & Ordinary averaging (\%) & \sfgl (\%) & Contribution (points)\\
\midrule
\multicolumn{5}{@{}l}{\emph{Additional WinoGrande items, FLASH replay}}\\
Corrected (base wrong, reference right) & 18.4 & 50.3 & 75.8 & $+$4.69 {\scriptsize[3.84, 5.55]}\\
Broken (base right, reference wrong) & 6.7 & 64.6 & 43.6 & $-$1.39 {\scriptsize[$-$1.61, $-$1.18]}\\
Both right & 58.4 & 96.1 & 95.2 & $-$0.48 {\scriptsize[$-$0.84, $-$0.12]}\\
Both wrong & 16.6 & 7.2 & 15.8 & $+$1.41 {\scriptsize[1.04, 1.77]}\\
\bottomrule
\end{tabular}
\caption{Qwen2.5-7B under FLASH replay: the item groups of \cref{tab:qwen-items} on the 767 additional WinoGrande items. Contributions add to the difference of 4.23 points.}
\label{tab:qwen-heldout-items}
\end{table*}

\subsection{General language modelling}\label{app:qwen-retention}
\begin{figure}[h]
\centering
\includegraphics[width=\linewidth]{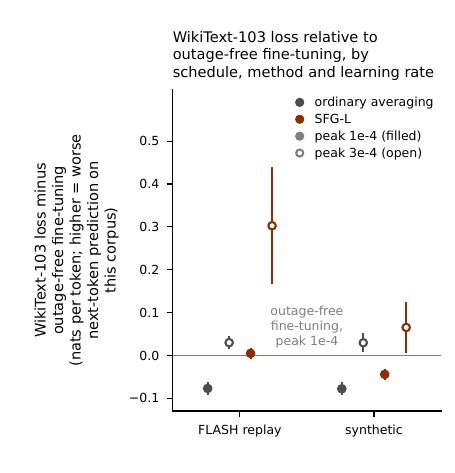}
\caption{Qwen2.5-7B: WikiText-103 loss of each rule's final client adapters minus that of outage-free fine-tuning at $10^{-4}$, six seeds, 95\% intervals. Higher means worse next-token prediction on this corpus. Both rules are shown at both peak learning rates; outage-free fine-tuning at $3\times10^{-4}$, not shown, is 1.22 nats above the reference.}
\label{fig:wikitext}
\end{figure}

The test split of WikiText-103 (raw version) is concatenated and tokenised, which gives 1,167 non-overlapping windows of 256 tokens; 200 of them are drawn once with a fixed seed and used for every run. For each of the 100 final client adapters, WikiText-103 loss is the mean next-token negative log-likelihood over the 51,000 predictions in those windows, with no prompt and no answer format. \Cref{tab:qwen-retention} gives the difference between the rules in the two selected configurations by seed set and the levels, and \cref{fig:wikitext} each rule relative to outage-free fine-tuning. In the selected configurations the differences have opposite signs, in all six seeds each. The negative sign under the synthetic schedule is a consequence of the baseline's higher learning rate: with both rules at $10^{-4}$ the difference there is $+0.034$ nats (0.021 to 0.047), and ordinary averaging's loss is 0.107 nats higher at $3\times10^{-4}$ than at $10^{-4}$ (\cref{app:qwen-shared}). For the first three seeds the stored values were used; re-evaluating the saved adapters of the three additional seeds, and of one earlier run, reproduced the stored values exactly.

\begin{table}[h]
\centering\small\setlength{\tabcolsep}{2.5pt}
\begin{tabular}{@{}lrr@{}}
\toprule
 & FLASH replay & Synthetic\\
\midrule
\multicolumn{3}{@{}l}{\emph{\sfgl minus ordinary averaging}}\\
Seeds 1--3 & $+$0.084 {\scriptsize[0.031, 0.137]} & $-$0.068 {\scriptsize[$-$0.112, $-$0.023]}\\
Seeds 4--6 & $+$0.080 {\scriptsize[0.033, 0.127]} & $-$0.080 {\scriptsize[$-$0.147, $-$0.013]}\\
All six & $+$0.082 {\scriptsize[0.063, 0.101]} & $-$0.074 {\scriptsize[$-$0.096, $-$0.051]}\\
\midrule
\multicolumn{3}{@{}l}{\emph{Level, six seeds}}\\
Ord.\ avg.\ & 2.349 {\scriptsize[2.348, 2.350]} & 2.455 {\scriptsize[2.422, 2.489]}\\
\sfgl & 2.431 {\scriptsize[2.411, 2.451]} & 2.382 {\scriptsize[2.367, 2.396]}\\
\bottomrule
\end{tabular}
\caption{Qwen2.5-7B, selected configurations: WikiText-103 loss of the final client adapters, nats per token, with 95\% intervals over seeds. Base model: 2.321; outage-free fine-tuning: 2.426.}
\label{tab:qwen-retention}
\end{table}

\subsection{The learning-rate grid}\label{app:qwen-shared}
\paragraph{Runs.} The first six-seed runs used the schedules selected in \cref{app:qwen-setup}, so they covered ordinary averaging and \sfgl at $10^{-4}$ under FLASH replay, and ordinary averaging at $3\times10^{-4}$ with \sfgl at $10^{-4}$ under the synthetic schedule. Two sets of runs on the same six seeds, with the data, graph, availability traces and evaluation unchanged, complete the grid of two rules, two peak rates and two schedules. The first added ordinary averaging at $10^{-4}$ under the synthetic schedule; the reading of the contrast between schedules was fixed before these runs, namely that the contrast persists at a shared rate if its 95\% interval excludes zero. The second added both rules at $3\times10^{-4}$ under FLASH replay and \sfgl at $3\times10^{-4}$ under the synthetic schedule. Its primary comparison, \sfgl at $10^{-4}$ against ordinary averaging at $3\times10^{-4}$ under FLASH replay on the common test, and its reading were fixed before the runs: an advantage over the higher-rate baseline is shown if the 95\% interval lies above zero, reversed if it lies below, and not shown otherwise. The other comparisons of the grid were listed beforehand as secondary, and the two selection criteria compared below as descriptive. The 12 new runs that repeat stored selection runs of the first three seeds reproduce them exactly.

\paragraph{Levels.} \Cref{tab:qwen-lr-levels} gives the levels of all ten conditions. Under outages every rule and schedule is more accurate at the higher rate, on WinoGrande and on the other tasks, and has higher WikiText-103 loss. Outage-free fine-tuning is not more accurate at the higher rate, and its WikiText-103 loss rises from 2.426 to 3.643 nats. Outage-free fine-tuning at $10^{-4}$ is therefore a reference and not a ceiling: under the synthetic schedule \sfgl at $3\times10^{-4}$ ends 1.09 points (0.71 to 1.48) above it on the common test.

\begin{table*}[t]
\centering\small
\begin{tabular}{@{}lrrrrr@{}}
\toprule
 & Common test & Own mixture & WinoGrande & Seven other tasks & WikiText-103 loss\\
\midrule
\multicolumn{6}{@{}l}{\emph{FLASH replay}}\\
\quad Ordinary averaging, $10^{-4}$ & 86.13 & 86.73 & 72.1 & 88.13 & 2.349\\
\quad Ordinary averaging, $3\times10^{-4}$ & 86.62 & 87.24 & 74.9 & 88.30 & 2.456\\
\quad \sfgl, $10^{-4}$ & 86.84 & 87.26 & 76.7 & 88.29 & 2.431\\
\quad \sfgl, $3\times10^{-4}$ & 87.25 & 87.48 & 78.3 & 88.52 & 2.728\\
\multicolumn{6}{@{}l}{\emph{Synthetic}}\\
\quad Ordinary averaging, $10^{-4}$ & 86.85 & 87.47 & 74.5 & 88.61 & 2.348\\
\quad Ordinary averaging, $3\times10^{-4}$ & 87.50 & 88.02 & 78.0 & 88.87 & 2.455\\
\quad \sfgl, $10^{-4}$ & 87.59 & 88.02 & 77.9 & 88.98 & 2.382\\
\quad \sfgl, $3\times10^{-4}$ & 88.44 & 88.57 & 80.1 & 89.63 & 2.491\\
\multicolumn{6}{@{}l}{\emph{Outage-free fine-tuning}}\\
\quad Ordinary averaging, $10^{-4}$ & 87.35 & 87.93 & 78.4 & 88.62 & 2.426\\
\quad Ordinary averaging, $3\times10^{-4}$ & 87.15 & 87.39 & 78.4 & 88.40 & 3.643\\
\bottomrule
\end{tabular}
\caption{Qwen2.5-7B, six seeds: levels for every rule, peak learning rate and availability schedule. Accuracies are in \%, averaged over clients; WikiText-103 loss is in nats per token (base model 2.321).}
\label{tab:qwen-lr-levels}
\end{table*}

\begin{figure*}[t]
\centering
\includegraphics[width=\textwidth]{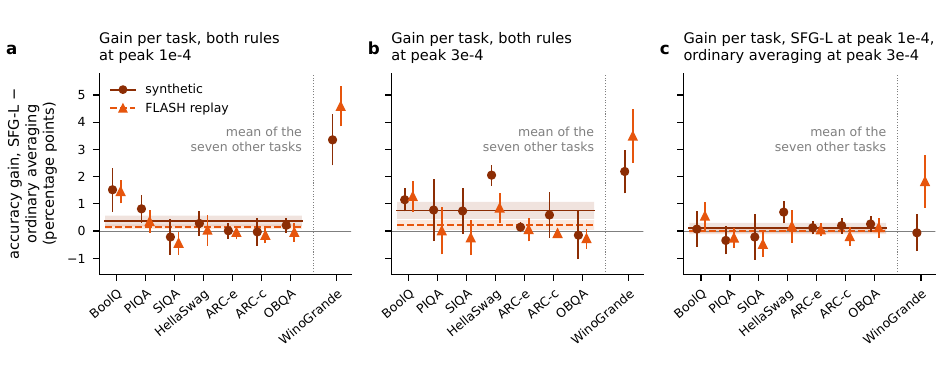}
\caption{Qwen2.5-7B task-level accuracy gains for \sfgl over ordinary averaging under both availability schedules. Both rules use (a)~$10^{-4}$ or (b)~$3\times10^{-4}$. In (c)~\sfgl uses $10^{-4}$ and ordinary averaging uses $3\times10^{-4}$. Lines and bands show the mean gain for the seven tasks other than WinoGrande and its 95\% interval. Error bars give 95\% intervals over six seeds.}
\label{fig:tasks-rates}
\end{figure*}

\paragraph{Comparisons.} \Cref{tab:qwen} gives the differences between the rules, and \cref{tab:qwen-shared-tasks} and \cref{fig:tasks-rates} the differences per task. The gain at matched rates is positive in all six seeds in each of the four cells. The contrast between FLASH replay and the synthetic schedule is $-$0.03 points ($-$0.28 to 0.23) at $10^{-4}$, so we did not detect the contrast at this shared rate under the prespecified criterion; at $3\times10^{-4}$ it is $-$0.30 ($-$0.76 to 0.15). Between the selected configurations it was 0.62 points. For the primary comparison of the second set, the difference of 0.22 points (0.09 to 0.36) lies above zero, so an advantage over the higher-rate baseline is shown under FLASH replay; on WinoGrande that difference is 1.82 points and on the seven other tasks $-$0.01. Per task, WinoGrande and BoolQ gain in all four matched comparisons, and HellaSwag in addition under the synthetic schedule at $3\times10^{-4}$, where WinoGrande contributes 29\% of the eight-task gain.

\begin{table*}[t]
\centering\scriptsize\setlength{\tabcolsep}{3pt}
\begin{tabular}{@{}lrrrrrr@{}}
\toprule
 & \multicolumn{2}{c}{Both rules at $10^{-4}$} & \multicolumn{2}{c}{Both rules at $3\times10^{-4}$} & \multicolumn{2}{c}{Ord.\ avg.\ $3\times10^{-4}$, \sfgl $10^{-4}$}\\
\cmidrule(lr){2-3}\cmidrule(lr){4-5}\cmidrule(lr){6-7}
Task & FLASH replay & Synthetic & FLASH replay & Synthetic & FLASH replay & Synthetic\\
\midrule
BoolQ & $+$1.45$^{\dagger}$ {\scriptsize[1.02, 1.89]} & $+$1.52$^{\dagger}$ {\scriptsize[0.71, 2.33]} & $+$1.27$^{\dagger}$ {\scriptsize[0.71, 1.83]} & $+$1.15$^{\dagger}$ {\scriptsize[0.73, 1.58]} & $+$0.55 {\scriptsize[0.02, 1.08]} & $+$0.07 {\scriptsize[$-$0.58, 0.73]}\\
PIQA & $+$0.35 {\scriptsize[$-$0.06, 0.76]} & $+$0.82 {\scriptsize[0.30, 1.33]} & $+$0.01 {\scriptsize[$-$0.86, 0.88]} & $+$0.77 {\scriptsize[$-$0.37, 1.92]} & $-$0.25 {\scriptsize[$-$0.61, 0.11]} & $-$0.34 {\scriptsize[$-$0.86, 0.17]}\\
Social IQa & $-$0.45 {\scriptsize[$-$0.87, $-$0.02]} & $-$0.22 {\scriptsize[$-$0.87, 0.43]} & $-$0.25 {\scriptsize[$-$0.89, 0.39]} & $+$0.74 {\scriptsize[$-$0.10, 1.57]} & $-$0.49 {\scriptsize[$-$0.94, $-$0.03]} & $-$0.22 {\scriptsize[$-$1.05, 0.61]}\\
HellaSwag & $+$0.03 {\scriptsize[$-$0.54, 0.61]} & $+$0.28 {\scriptsize[$-$0.18, 0.74]} & $+$0.85 {\scriptsize[0.29, 1.41]} & $+$2.06$^{\dagger}$ {\scriptsize[1.67, 2.44]} & $+$0.18 {\scriptsize[$-$0.43, 0.79]} & $+$0.70 {\scriptsize[0.28, 1.11]}\\
WinoGrande & $+$4.58$^{\dagger}$ {\scriptsize[3.85, 5.32]} & $+$3.35$^{\dagger}$ {\scriptsize[2.42, 4.29]} & $+$3.49$^{\dagger}$ {\scriptsize[2.51, 4.47]} & $+$2.19$^{\dagger}$ {\scriptsize[1.39, 2.99]} & $+$1.82$^{\dagger}$ {\scriptsize[0.86, 2.78]} & $-$0.06 {\scriptsize[$-$0.74, 0.61]}\\
ARC-Easy & $-$0.05 {\scriptsize[$-$0.30, 0.20]} & $+$0.01 {\scriptsize[$-$0.28, 0.31]} & $+$0.06 {\scriptsize[$-$0.38, 0.49]} & $+$0.14 {\scriptsize[$-$0.04, 0.33]} & $+$0.05 {\scriptsize[$-$0.19, 0.29]} & $+$0.12 {\scriptsize[$-$0.12, 0.37]}\\
ARC-Challenge & $-$0.16 {\scriptsize[$-$0.44, 0.13]} & $-$0.03 {\scriptsize[$-$0.54, 0.48]} & $-$0.09 {\scriptsize[$-$0.22, 0.04]} & $+$0.59 {\scriptsize[$-$0.25, 1.43]} & $-$0.20 {\scriptsize[$-$0.53, 0.14]} & $+$0.20 {\scriptsize[$-$0.10, 0.50]}\\
OpenBookQA & $-$0.05 {\scriptsize[$-$0.39, 0.30]} & $+$0.21 {\scriptsize[$-$0.07, 0.49]} & $-$0.27 {\scriptsize[$-$0.64, 0.09]} & $-$0.15 {\scriptsize[$-$1.03, 0.73]} & $+$0.13 {\scriptsize[$-$0.24, 0.49]} & $+$0.26 {\scriptsize[$-$0.03, 0.56]}\\
\midrule
Seven other tasks & $+$0.16 {\scriptsize[0.08, 0.25]} & $+$0.37 {\scriptsize[0.18, 0.56]} & $+$0.23 {\scriptsize[0.04, 0.41]} & $+$0.76 {\scriptsize[0.44, 1.07]} & $-$0.01 {\scriptsize[$-$0.11, 0.10]} & $+$0.11 {\scriptsize[$-$0.08, 0.31]}\\
Common test & $+$0.72 {\scriptsize[0.62, 0.81]} & $+$0.74 {\scriptsize[0.54, 0.94]} & $+$0.63 {\scriptsize[0.45, 0.82]} & $+$0.94 {\scriptsize[0.61, 1.26]} & $+$0.22 {\scriptsize[0.09, 0.36]} & $+$0.09 {\scriptsize[$-$0.12, 0.30]}\\
\bottomrule
\end{tabular}
\caption{Qwen2.5-7B, six seeds: \sfgl minus ordinary averaging per task on the common test, in percentage points with 95\% intervals, for the six combinations of peak learning rates and availability schedule. $^{\dagger}$The interval excludes zero after Holm correction over the eight tasks of that column.}
\label{tab:qwen-shared-tasks}
\end{table*}

\paragraph{Selection criteria.} \Cref{tab:qwen-selection} gives the difference in validation accuracy between the two rates under two criteria. The criterion fixed before the experiment, validation accuracy averaged over training, selects $10^{-4}$ for both rules under FLASH replay, by 0.04 points for ordinary averaging, and $3\times10^{-4}$ for ordinary averaging only under the synthetic schedule. Validation accuracy at the final round selects $3\times10^{-4}$ for both rules under both schedules, as does the common-test accuracy itself. A criterion averaged over training favours a rate that learns faster early; the two criteria therefore answer different questions, and the comparison at separately selected rates inherits that choice. We report the grid in full for this reason.

\begin{table}[h]
\centering\small\setlength{\tabcolsep}{3pt}
\begin{tabular}{@{}lrrrr@{}}
\toprule
 & \multicolumn{2}{c}{FLASH replay} & \multicolumn{2}{c}{Synthetic}\\
\cmidrule(lr){2-3}\cmidrule(lr){4-5}
 & Ord.\ avg. & \sfgl & Ord.\ avg. & \sfgl\\
\midrule
\multicolumn{5}{@{}l}{\emph{Averaged over training}}\\
\quad Seeds 1--3 & $-$0.04 & $-$0.21 & $+$0.11 & $-$0.40\\
\quad All six seeds & $-$0.08 & $-$0.29 & $+$0.07 & $-$0.33\\
\multicolumn{5}{@{}l}{\emph{Final round}}\\
\quad Seeds 1--3 & $+$0.45 & $+$0.16 & $+$0.45 & $+$0.23\\
\quad All six seeds & $+$0.40 & $+$0.10 & $+$0.40 & $+$0.28\\
\bottomrule
\end{tabular}
\caption{Qwen2.5-7B: validation accuracy at a peak rate of $3\times10^{-4}$ minus that at $10^{-4}$, in percentage points, under two selection criteria. A positive value selects the higher rate. Selection used the first three seeds and the criterion averaged over training.}
\label{tab:qwen-selection}
\end{table}

\subsection{Qwen2.5-1.5B}\label{app:qwen-small}
We also ran the smaller Qwen2.5-1.5B in the same setting. The endpoint of those runs is accuracy on clients' own task mixtures; they have no common-test evaluation. Over five seeds and both availability schedules the difference is 0.38 points (0.13 to 0.64). The interval excludes zero and includes the 0.5-point threshold, so the result is unresolved in the same sense as the pooled 7B estimate.

\subsection{Availability schedules}\label{app:traces}
\paragraph{FLASH replay.} We use the state trace of 1,000 phones published with FLASH \citep{yang2021flash}. Each phone's events are filled forward into a per-minute state, and a phone counts as available in a minute when its screen is off, it is charging and it is on Wi-Fi. A round represents ten minutes and is available if the phone is available in at least five of them. Each phone's window of 300 rounds starts at the first midnight after its first event. Phones without a full window are dropped (797 remain), as are phones available less than 5\% of the time, which leaves 326. Each seed draws 100 of them without replacement, one per client, and both rules of a pair share the draw.

\paragraph{Synthetic schedule.} Each client follows an independent two-state Markov chain with 80\% of rounds offline and a mean offline spell of 10 rounds.

\paragraph{Diagnostics.} \Cref{tab:traces} compares the realised schedules in the 100-client setting. The two have nearly the same share of rounds offline. FLASH replay has offline and online spells almost five times as long, twice the variation in how many clients are available together, three times the variation between clients in how often they are available, and fewer clients available in the last third of a run.

\begin{table}[h]
\centering\footnotesize\setlength{\tabcolsep}{3pt}
\begin{tabular}{@{}>{\raggedright\arraybackslash}p{0.62\linewidth}rr@{}}
\toprule
 & FLASH & Synthetic\\
 & replay & \\
\midrule
Share of rounds offline & 0.82 & 0.80\\
Mean offline spell (rounds) & 46.2 & 9.6\\
Median offline spell (rounds) & 31.8 & 7.0\\
90th percentile of offline spell & 108 & 21\\
Share of offline spells of 50 rounds or more & 0.35 & 0.00\\
Mean online spell (rounds) & 11.5 & 2.5\\
Variance of clients available per round (independent clients = 1) & 2.11 & 1.02\\
Std.\ dev.\ across clients of share of rounds available & 0.12 & 0.04\\
Share of clients available, rounds 1--100 & 0.199 & 0.195\\
Share of clients available, rounds 101--200 & 0.197 & 0.202\\
Share of clients available, rounds 201--300 & 0.145 & 0.200\\
\bottomrule
\end{tabular}
\caption{Realised availability schedules for 100 clients and 300 rounds, means over the six seeds.}
\label{tab:traces}
\end{table}

\paragraph{Three further schedules.} The schedules of \cref{sec:qwen-schedule} are built as follows. \emph{Shifted per client}: each client's FLASH trace is rotated in time by its own offset, drawn uniformly at random; a client keeps its sequence of offline and online spells, and clients no longer share a time profile. \emph{Shifted by a common offset}: all clients' traces are rotated by the same offset, so the co-availability of clients is that of FLASH replay and only its placement within training changes. \emph{Synthetic, matched to FLASH}: independent two-state Markov chains with the share of rounds offline and the mean offline spell of FLASH replay; clients are independent and statistically identical. Both rules ran at a peak rate of $10^{-4}$ on six seeds, paired with the FLASH replay runs of the same seeds. The design, the three contrasts and a margin of 0.25 points for calling two gains equivalent were fixed before the runs. As a check, two stored runs were first repeated and reproduced exactly.

\begin{figure*}[t]
\centering
\includegraphics[width=\textwidth]{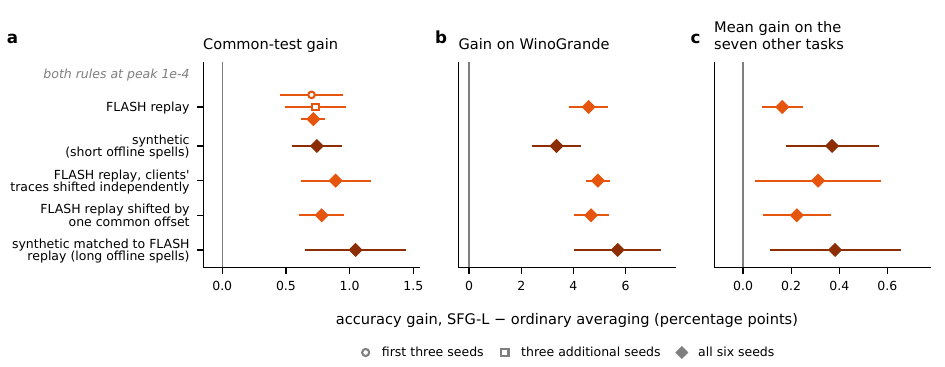}
\caption{Qwen2.5-7B accuracy gains for \sfgl over ordinary averaging across five availability schedules, with both rules at $10^{-4}$. Panels show (a)~the common test, (b)~WinoGrande and (c)~the mean of the seven other tasks. For FLASH replay, panel (a) shows the original three seeds, the three additional seeds and all six. Error bars give 95\% intervals over the seeds in each estimate.}
\label{fig:schedules}
\end{figure*}

\paragraph{Result.} \Cref{tab:schedules-full} gives the diagnostics of each schedule and the results, and \cref{fig:schedules} the gains on the common test, WinoGrande and the mean of the seven other tasks. Per-client shifting removes the excess variance in the number of clients available together (0.88 against 2.11) and evens out availability over training. In the matched synthetic schedule clients are statistically identical, although with spells this long the realised shares of rounds available still differ between clients (0.10 against 0.12). The common-test gain is present on all five schedules, WinoGrande gains between 3.4 and 5.7 points on each, and \sfgl's WikiText-103 loss is higher than ordinary averaging's on each. The three contrasts fixed beforehand were inconclusive: FLASH replay minus the per-client shift is $-0.17$ points ($-0.52$ to 0.17); the per-client shift minus the matched synthetic schedule $-0.16$ ($-0.55$ to 0.23); the common offset minus the per-client shift $-0.11$ ($-0.43$ to 0.21). Each interval includes zero and extends beyond the margin, so neither a contribution of the property nor its absence is established.

\begin{table*}[t]
\centering\scriptsize\setlength{\tabcolsep}{2.5pt}
\begin{tabular}{@{}lrrrrr rrrr@{}}
\toprule
 & \multicolumn{5}{c}{Schedule diagnostics} & \multicolumn{4}{c}{\sfgl minus ordinary averaging}\\
\cmidrule(lr){2-6}\cmidrule(l){7-10}
Schedule & Offline & Spell & Var. & Clients & Late & Common test & WinoGrande & Other seven & WikiText\\
\midrule
FLASH replay & 0.82 & 46.2 & 2.11 & 0.12 & 0.145 & 0.72 {\scriptsize[0.62, 0.81]} & 4.58 {\scriptsize[3.85, 5.32]} & 0.16 {\scriptsize[0.08, 0.25]} & $+$0.082 {\scriptsize[0.063, 0.101]}\\
\quad shifted per client & 0.82 & 45.4 & 0.88 & 0.12 & 0.175 & 0.89 {\scriptsize[0.62, 1.16]} & 4.94 {\scriptsize[4.49, 5.38]} & 0.31 {\scriptsize[0.05, 0.57]} & $+$0.122 {\scriptsize[0.086, 0.158]}\\
\quad shifted by a common offset & 0.82 & 45.1 & 2.11 & 0.12 & 0.186 & 0.78 {\scriptsize[0.60, 0.96]} & 4.68 {\scriptsize[4.01, 5.34]} & 0.22 {\scriptsize[0.08, 0.36]} & $+$0.093 {\scriptsize[0.066, 0.119]}\\
Synthetic, matched to FLASH & 0.82 & 45.6 & 1.07 & 0.10 & 0.180 & 1.05 {\scriptsize[0.65, 1.44]} & 5.69 {\scriptsize[4.02, 7.37]} & 0.38 {\scriptsize[0.11, 0.65]} & $+$0.088 {\scriptsize[0.070, 0.106]}\\
Synthetic, short spells & 0.80 & 9.6 & 1.02 & 0.04 & 0.200 & 0.74 {\scriptsize[0.54, 0.94]} & 3.35 {\scriptsize[2.42, 4.29]} & 0.37 {\scriptsize[0.18, 0.56]} & $+$0.034 {\scriptsize[0.021, 0.047]}\\
\bottomrule
\end{tabular}
\caption{Qwen2.5-7B, both rules at a peak rate of $10^{-4}$, six seeds. \emph{Offline}: share of rounds offline. \emph{Spell}: mean offline spell (rounds). \emph{Var.}: variance of the number of clients available per round, relative to independent clients. \emph{Clients}: standard deviation across clients of the share of rounds available. \emph{Late}: share of clients available in rounds 201--300. Differences are in percentage points of accuracy, and in nats per token for WikiText-103 loss, with 95\% intervals.}
\label{tab:schedules-full}
\end{table*}

\section{\texorpdfstring{GPT-124M pretraining: details for \cref{sec:gpt}}{GPT-124M pretraining details}}\label{app:gpt}

\subsection{Setup}\label{app:gpt-setup}
\paragraph{Model, data and optimiser.} GPT-124M is the GPT-2-small architecture \citep{radford2019gpt2} with 124,373,760 parameters, trained on FineWeb-Edu \citep{penedo2024fineweb}. 16 clients sit on a 3-nearest-neighbour graph and hold identically distributed data. A run is 300 rounds; in each round an online client takes 50 AdamW steps \citep{loshchilov2019adamw} on batches of $16\times1024$ tokens, with weight decay 0.1, gradient clipping at 1.0 and a warm-up of 100 steps per client. Learning rates follow a cosine schedule over the run. Clients exchange full models.

\paragraph{Measurement.} Validation uses 512 held-out sequences of 1024 tokens. We compute each client's validation loss, average over the 16 clients, and report $100\,(1-\mathrm{ppl}_{\text{\sfgl}}/\mathrm{ppl}_{\text{ordinary averaging}})$, the percentage by which \sfgl's perplexity is lower. All cost quantities below use the same form, $100\,(1-\mathrm{ppl}_a/\mathrm{ppl}_b)$ for two training conditions $a$ and $b$.

\paragraph{Availability schedules.} The schedules are those of \cref{app:traces}. With 16 clients, the phones assigned from the FLASH trace are close to independent of one another, unlike in the 100-client setting. The stored trace covers 300 rounds. In the 900-round runs, every round from the 300th on replays the availability of 288 rounds earlier, which is exactly two days, so the time of day is preserved and the trace is repeated, not extended with new recordings.

\paragraph{Choosing learning rates.} For every start, the peak learning rate of each rule was chosen on seeds that were not used for the comparison, by the loss on a separate selection set averaged over the training run. This criterion can prefer a lower rate than the final loss alone would. From random initialisation the candidates were $6\times10^{-4}$ and $1.2\times10^{-3}$ (with $2.4\times10^{-3}$ added in \cref{app:gpt-offline}); ordinary averaging selected $1.2\times10^{-3}$ and \sfgl $6\times10^{-4}$, on two seeds, and the comparison used three others. The same three comparison seeds serve the 900-round runs, the original maturity experiment and the matched-work reference. The repeat of \cref{app:gpt-repeat} used nine further seeds: three to select rates and six to compare.

\subsection{Training duration}\label{app:gpt-long}
\Cref{tab:gpt-long} gives the reduction at three points of the 900-round runs. Both rules ran at a peak rate of $6\times10^{-4}$ with the cosine schedule stretched over 900 rounds, so the value after the first third is not that of the original 300-round experiment, in which each rule ran at its own rate and the schedule had already decayed. The advantage also holds on text from another source: on WikiText-103 it is 15.5\% at the end. Comparing the best client of each rule instead of the client average gives 12.5\%. We do not interpret the model obtained by averaging all clients' parameters in these runs: from random initialisation that average was unstable, and the estimated reduction for it is $-$96 [$-$353, 161]\%.

\begin{table}[h]
\centering\small
\begin{tabular}{@{}lr@{}}
\toprule
Point of the run & Reduction (\%)\\
\midrule
After 300 rounds & 19.0 {\scriptsize[15.1, 22.9]}\\
After 600 rounds & 14.7 {\scriptsize[12.7, 16.8]}\\
After 900 rounds & 13.5 {\scriptsize[11.4, 15.6]}\\
\quad synthetic only & 15.0 {\scriptsize[14.4, 15.5]}\\
\quad FLASH replay only & 12.1 {\scriptsize[7.4, 16.8]}\\
After 900 rounds, WikiText-103 & 15.5 {\scriptsize[13.2, 17.8]}\\
After 900 rounds, best client & 12.5 {\scriptsize[9.5, 15.4]}\\
\bottomrule
\end{tabular}
\caption{GPT-124M from random initialisation, 900 rounds, both rules at the same learning rate: \% lower validation perplexity with \sfgl. Six seed-and-schedule pairs, or the three seeds of one availability schedule; 95\% intervals. The headline value in \cref{sec:gpt} uses seeds as the unit (13.5\%, 11.4 to 15.6).}
\label{tab:gpt-long}
\end{table}

\subsection{Starting maturity: the original three seeds}\label{app:gpt-maturity}
\Cref{tab:gpt-maturity} gives the gain by start and availability schedule, with each rule at its selected learning rate. At the two most mature starts the rules selected different rates under the synthetic schedule and the same rate under FLASH replay. The pattern of the later repeat is already visible: no gain under the synthetic schedule, and 3.7\% under FLASH replay.

\begin{table*}[t]
\centering\footnotesize\setlength{\tabcolsep}{4pt}
\begin{tabular}{@{}llrrr@{}}
\toprule
Start & Peak rate, ord.\ avg.\ / \sfgl & Both schedules & Synthetic & FLASH replay\\
\midrule
Random initialisation & $1.2\times10^{-3}$ / $6\times10^{-4}$ & 22.0 [15.6, 28.4] & 26.3 [25.3, 27.2] & 17.7 [5.0, 30.4]\\
0.1B tokens & $3\times10^{-4}$ / $3\times10^{-4}$ & 8.4 [3.2, 13.6] & 8.9 [6.3, 11.4] & 8.0 [0.0, 15.9]\\
0.5B tokens & $3\times10^{-4}$, $1.5\times10^{-4}$ / $1.5\times10^{-4}$ & 1.9 [$-$0.2, 4.1] & 0.1 [$-$0.9, 1.2] & 3.7 [0.4, 7.0]\\
2.5B tokens & $3\times10^{-4}$, $1.5\times10^{-4}$ / $1.5\times10^{-4}$ & 1.8 [$-$0.5, 4.2] & 0.0 [$-$1.2, 1.2] & 3.7 [0.2, 7.2]\\
\bottomrule
\end{tabular}
\caption{GPT-124M, starting maturity, original three seeds: \% lower validation perplexity with \sfgl, each rule at its selected peak learning rate (where two rates are given for ordinary averaging, the first is for the synthetic schedule and the second for FLASH replay). 95\% intervals over seeds.}
\label{tab:gpt-maturity}
\end{table*}

\begin{table*}[!t]
\centering\small
\begin{tabular}{@{}llrrr@{}}
\toprule
Start & Schedule & Both at baseline's rate & Each at its own rate & WikiText-103, baseline's rate\\
\midrule
0.5B tokens & synthetic & 2.2 [1.7, 2.7] & $-$0.2 [$-$0.6, 0.3] & 2.3 [1.6, 3.0]\\
 & FLASH replay & 4.2 [3.1, 5.3] & 4.2 [3.1, 5.3] & 4.2 [3.2, 5.2]\\
 & both & 3.2 [2.6, 3.8] & 2.0 [1.4, 2.6] & 3.3 [2.9, 3.6]\\
2.5B tokens & synthetic & 2.0 [1.6, 2.4] & $-$0.2 [$-$0.7, 0.2] & 2.8 [2.1, 3.5]\\
 & FLASH replay & 4.1 [3.0, 5.2] & 4.1 [3.0, 5.2] & 5.7 [4.3, 7.0]\\
 & both & 3.0 [2.5, 3.6] & 1.9 [1.4, 2.5] & 4.2 [3.6, 4.9]\\
\bottomrule
\end{tabular}
\caption{GPT-124M, repeat on six new seeds: \% lower perplexity with \sfgl; 95\% intervals over seeds. Under FLASH replay the two rules selected the same rate, so the first two columns coincide.}
\label{tab:gpt-repeat}
\end{table*}

\begin{table*}[!t]
\centering\small
\begin{tabular}{@{}llrrrrr@{}}
\toprule
Start & Schedule & $7.5\times10^{-5}$ & $1.5\times10^{-4}$ & $3\times10^{-4}$ & $6\times10^{-4}$ & $1.2\times10^{-3}$\\
\midrule
0.5B tokens & synthetic & 3.7 & 3.0 & 1.8 & $-$0.2 & $-$2.9\\
 & FLASH replay & 3.9 & 3.1 & 1.9 & $-$0.3 & $-$2.1\\
2.5B tokens & synthetic & 3.7 & 3.0 & 1.6 & $-$1.0 & $-$7.8\\
 & FLASH replay & 3.8 & 3.0 & 1.5 & $-$1.3 & $-$8.1\\
\bottomrule
\end{tabular}
\caption{GPT-124M, selection seeds of the repeat (three seeds): \% lower validation perplexity with \sfgl when both rules use the same peak learning rate.}
\label{tab:gpt-tune}
\end{table*}

\subsection{Starting maturity: the repeat on six new seeds}\label{app:gpt-repeat}
Because three seeds could not separate a small gain from none, we repeated the two most mature starts. The design was fixed before the runs: five candidate peak rates from $7.5\times10^{-5}$ to $1.2\times10^{-3}$; rates selected for each rule, start and availability schedule on three seeds; the comparison on six further seeds; and the primary comparison with both rules at the rate selected for ordinary averaging. The possible readings were also fixed: a \emph{gain} (above zero after correction for the comparisons made, with a 95\% interval reaching 2\%), \emph{no practically relevant gain} (equivalent to zero within $\pm2$\%), a \emph{reversal}, or \emph{inconclusive}. Both starts were read as a gain with both rules at the baseline's rate; under the synthetic schedule with each rule at its own rate the reading was no practically relevant gain. Ordinary averaging selected $3\times10^{-4}$ under the synthetic schedule and $1.5\times10^{-4}$ under FLASH replay; \sfgl selected $1.5\times10^{-4}$ throughout. No selection lay at the edge of the candidate range. All runs completed.

\Cref{tab:gpt-repeat} gives the results and \cref{tab:gpt-tune} the gain at each shared rate on the selection seeds. Three points qualify the headline of \cref{sec:gpt}. First, with each rule at its own rate the gain under the synthetic schedule is not detected at either start. Second, \sfgl's selected rate was not the rate at which it ended best: under the synthetic schedule \sfgl at ordinary averaging's rate ended better than \sfgl at its own, which is why the common-rate gain exceeds the own-rate gain. Third, the gain at a shared rate falls from about 3.8\% at the lowest rate to a loss at the two highest, at both starts and under both availability schedules. A reader who tunes both rules by final loss should therefore expect a smaller gain under the synthetic schedule than our primary comparison shows.

\begin{table*}[t]
\centering\small
\begin{tabular}{@{}llrrrr@{}}
\toprule
Start & Schedule & Outage cost & Lost work & Lost communication & \sfgl gain\\
\midrule
\multicolumn{6}{@{}l}{\emph{Original three seeds, each rule at its own rate}}\\
Random initialisation & both & 55.4 [48.0, 62.8] & 43.3 [38.7, 48.0] & 21.7 [12.7, 30.7] & 22.0 [16.0, 28.0]\\
0.1B tokens & both & 26.1 [22.9, 29.3] & 23.6 [21.8, 25.3] & 3.3 [1.0, 5.7] & 8.4 [6.1, 10.7]\\
0.5B tokens & both & 13.6 [10.1, 17.2] & 12.2 [9.5, 14.9] & 1.6 [0.6, 2.6] & 1.9 [$-$0.3, 4.2]\\
2.5B tokens & both & 12.8 [9.2, 16.5] & 11.4 [8.6, 14.2] & 1.6 [0.6, 2.6] & 1.8 [$-$0.5, 4.2]\\
\midrule
\multicolumn{6}{@{}l}{\emph{Six new seeds, both rules at ordinary averaging's rate}}\\
0.5B tokens & synthetic & 10.5 [10.4, 10.5] & 9.9 [9.8, 10.0] & 0.6 [0.5, 0.7] & 2.2 [1.7, 2.7]\\
 & FLASH replay & 11.4 [10.4, 12.5] & 9.3 [8.4, 10.3] & 2.3 [2.0, 2.6] & 4.2 [3.1, 5.3]\\
 & both & 11.0 [10.4, 11.5] & 9.6 [9.1, 10.1] & 1.5 [1.3, 1.6] & 3.2 [2.6, 3.8]\\
2.5B tokens & synthetic & 9.6 [9.5, 9.7] & 9.0 [8.9, 9.2] & 0.6 [0.5, 0.8] & 2.0 [1.6, 2.4]\\
 & FLASH replay & 11.2 [10.2, 12.3] & 9.2 [8.2, 10.1] & 2.3 [1.9, 2.6] & 4.1 [3.0, 5.2]\\
 & both & 10.4 [9.9, 10.9] & 9.1 [8.6, 9.6] & 1.4 [1.3, 1.6] & 3.0 [2.5, 3.6]\\
\bottomrule
\end{tabular}
\caption{GPT-124M: outage cost of ordinary averaging and its two parts, and the gain of \sfgl, all as \% lower validation perplexity (definitions in the text). Original seeds: intervals over the six seed-and-schedule pairs. New seeds: intervals over seeds.}
\label{tab:gpt-split}
\end{table*}

\begin{table*}[t]
\centering\small
\begin{tabular}{@{}lrrrr@{}}
\toprule
Rounds offline & \sfgl gain (\%) & Cost of outages (\%) & Fraction recovered & \sfgl gain, WikiText-103 (\%)\\
\midrule
20\% & 3.6 [2.7, 4.5] & 6.1 [4.4, 7.8] & 0.60 [0.48, 0.72] & 4.7 [3.0, 6.3]\\
50\% & 13.6 [11.6, 15.6] & 20.2 [19.9, 20.6] & 0.67 [0.58, 0.76] & 15.0 [13.1, 16.9]\\
80\% & 26.3 [25.3, 27.2] & 51.6 [48.2, 54.9] & 0.51 [0.48, 0.54] & 31.9 [25.4, 38.3]\\
\bottomrule
\end{tabular}
\caption{GPT-124M from random initialisation under the synthetic schedule, by the share of rounds offline; three seeds, 95\% intervals. The fraction recovered is the gain divided by the outage cost, per seed.}
\label{tab:gpt-offline}
\end{table*}

\subsection{A reference with missed communication restored}\label{app:gpt-split}
The reference of \cref{sec:gpt} is ordinary averaging with the missed communication restored: every client exchanges and averages models in every round with equal weights, as if no client were ever out of reach, but only the clients that the schedule marks as available train. Its clients therefore do exactly the local work of the runs under outages with none of the missed exchanges. With $\mathrm{ppl}_{\text{none}}$, $\mathrm{ppl}_{\text{ref}}$ and $\mathrm{ppl}_{\text{D}}$ the perplexities of outage-free training, of the reference, and of ordinary averaging under outages, the outage cost is $100\,(1-\mathrm{ppl}_{\text{none}}/\mathrm{ppl}_{\text{D}})$, lost work is $100\,(1-\mathrm{ppl}_{\text{none}}/\mathrm{ppl}_{\text{ref}})$ and lost communication is $100\,(1-\mathrm{ppl}_{\text{ref}}/\mathrm{ppl}_{\text{D}})$. The reference and the outage-free run use the learning rate of ordinary averaging.

\Cref{tab:gpt-split} gives the split at all four starts on the original seeds and at the two mature starts on the new seeds. Before the runs we had set 2\% as the smallest cost of lost communication for which we would report the fraction that \sfgl recovers. From random initialisation the cost is 21.7\% and the gain is of the same size. At the 2.5-billion-token start the cost was below the threshold on the original seeds (1.6\%) and again on the new ones (1.4\%), so no fraction is reported at maturity.

\paragraph{Why no recovered fraction is reported at maturity.} A recovered fraction divides the gain by the cost of lost communication. At the mature starts that cost is between about 1\% and 2\% of perplexity, so the ratio has a small and noisy denominator: on the original seeds the mean ratio at the 2.5-billion-token start was 0.77 with an interval of $-$0.28 to 1.82. The 2\% threshold was set beforehand to avoid interpreting such ratios.

\paragraph{The reference is not a ceiling.} On the new seeds the gain exceeded the cost of lost communication in all 12 seed-and-schedule pairs at the 2.5-billion-token start, and \sfgl under outages ended 1.6\% below the reference with communication restored. The reference restores the missed exchanges but still averages with equal weights. A rule cannot execute local steps that were skipped, but it can compensate for part of their effect on performance, and here it did more than restoring the exchanges would have. We have not established why.

\subsection{How often clients are offline}\label{app:gpt-offline}
From random initialisation under the synthetic schedule, the gain grows with the share of rounds that clients spend offline (\cref{tab:gpt-offline}), and \sfgl recovers between half and two thirds of what outages cost ordinary averaging at every level. Each rule ran at a rate selected on two separate seeds; ordinary averaging's was selected from three candidates.

\section{A vision control: the baseline's learning rate and the cost of outages}\label{app:vision}
\Cref{sec:gpt} notes that the baseline's learning rate also changes how costly outages appear. The clearest case we have is outside language: a small convolutional network trained on CIFAR-10 \citep{krizhevsky2009cifar} by 100 clients with heterogeneous label distributions, starting from a checkpoint pretrained centrally for four epochs. The reference with communication restored is defined as in \cref{app:gpt-split} and uses the schedule of ordinary averaging. For the broader vision evidence on the rule we refer to \citet{wang2026sfg}. \Cref{tab:cnn-lr} gives the cost of lost communication, in points of accuracy on a class-balanced test set, when ordinary averaging's schedule is selected in two ways. Selection on validation data pooled over all classes chose much smaller learning rates than selection on clients' own validation data. At those rates clients barely move from the pretrained checkpoint, and missed exchanges then cost little.

\begin{table}[!htbp]
\centering\small
\begin{tabular}{@{}lrr@{}}
\toprule
Selected on & Synthetic & FLASH replay\\
\midrule
Clients' own data & 7.42 {\scriptsize[6.91, 7.93]} & 7.45 {\scriptsize[6.58, 8.33]}\\
Pooled data & 0.59 {\scriptsize[0.48, 0.70]} & 1.67 {\scriptsize[0.99, 2.35]}\\
\bottomrule
\end{tabular}
\caption{CNN on CIFAR-10 after four epochs of central pretraining: cost of lost communication for ordinary averaging (matched-work reference minus ordinary averaging under outages), in percentage points of accuracy on a class-balanced test set, with 95\% intervals.}
\label{tab:cnn-lr}
\end{table}

\section{Materials and compute}\label{app:repro}
\paragraph{Materials.} We are preparing the experiment code, configurations, and evaluation scripts for public release. The base models and datasets are public.

\paragraph{Compute.} All experiments ran on one server with A100 GPUs with 80\,GB of memory. A Qwen2.5-7B run of 300 rounds takes about 4.4 GPU-hours under outages and about 8 when every client is available in every round. The runs on the three further availability schedules (\cref{app:traces}) took 190 GPU-hours, the 24 runs that complete the learning-rate grid (\cref{app:qwen-shared}) 95, and the evaluations of saved adapters on additional items and other prompt formats 7. The repeat of the GPT-124M maturity experiment (\cref{app:gpt-repeat}) comprised 214 runs and 374 GPU-hours.

\end{document}